\documentclass[
    11pt,
]{amsart}

\pdfoutput=1

\usepackage[
    backend=biber,
    style=numeric,
    url=false,
]{biblatex}
\DeclareNameAlias{author}{family-given}
\usepackage[
    paper=a4paper, 
    margin=2cm,
]{geometry}
\usepackage{color}
\usepackage[
    dvipsnames,
]{xcolor}
\usepackage{csquotes}
\usepackage{xurl}
\usepackage{hyperref}
\hypersetup{
    hidelinks,
    colorlinks=true,
    allcolors=OliveGreen,
    breaklinks=true,
    addtopdfcreator=text,
}
\usepackage[
    capitalise, 
    nameinlink, 
    noabbrev, 
    nosort,
]{cleveref}
\usepackage{doi}
\usepackage[
    english,
]{babel}

\usepackage{booktabs}
\usepackage{pgfplots}
\pgfplotsset{
    width=10cm,
    compat=1.9
}
\usepackage{graphicx}
\usepackage{caption}
\usepackage{subcaption}
\usepackage{float}
\usepackage{tabularx}
\usepackage{ragged2e}
\newcolumntype{L}[1]{>{\hsize=#1\hsize\RaggedRight} X}

\newcommand{\LEE}{Liquid and electric energy}
\newcommand{\REM}{Emissions intensity from new vehicle registrations}
\newcommand{\ETV}{Emissions associated to national traffic volume}
\newcommand{\COA}{Weight of coal production}
\newcommand{\COW}{Number of dairy cattle}
\newcommand{\MEA}{Weight of meat produced at export-inspected facilities}
\newcommand{\FIM}{Quantity of imported fertilisers}
\newcommand{\EXM}{Food exports for meat}
\newcommand{\EXD}{Food exports for aquaculture and dairy}
\newcommand{\EXF}{Food exports for horticulture}

\hypersetup{
    pdfauthor={
        Malcolm Jones, 
        Hannah Chorley, 
        Flynn Owen, 
        Tamsyn Hilder, 
        Holly Trowland, 
        Paul Bracewell
    },
    pdftitle={
        Dynamic nowcast of the New Zealand greenhouse gas inventory
    }
}

\title[Dynamic nowcast of the New Zealand greenhouse gas inventory]{Dynamic nowcast of \\ the New Zealand greenhouse gas inventory}

\author[Jones]{Malcolm Jones}
\author[Chorley]{Hannah Chorley}
\author[Owen]{Flynn Owen}
\author[Hilder]{Tamsyn Hilder}
\author[Trowland]{Holly Trowland}
\author[Bracewell]{Paul Bracewell}

\address[M.\ Jones]{\textit{Corresponding author}, DOT Loves Data, Level One, 80 Willis Street, Wellington Central, 6011 New Zealand}
\email{\href{mailto:hello@malcolmjones.me}{hello@malcolmjones.me}}

\address[H.\ Chorley]{DOT Loves Data, Level One, 80 Willis Street, Wellington Central, 6011 New Zealand}
\email{\href{mailto:hkchorley@gmail.com}{hkchorley@gmail.com}}

\address[F.\ Owen, T.\ Hilder, H.\ Trowland, P.\ Bracewell]{DOT Loves Data, Level One, 80 Willis Street, Wellington Central, 6011 New Zealand}
\email{\href{mailto:flynn@dotlovesdata.com}{flynn}, \href{mailto:tamsyn@dotlovesdata.com}{tamsyn}, \href{mailto:holly@dotlovesdata.com}{holly}, \href{mailto:paul@dotlovesdata.com}{paul@dotlovesdata.com}}

\thanks{This research is supported by Transpower, Callaghan Innovation and DOT Loves Data.}


\date{\today.}

\begin{document}

\begin{abstract}
    As efforts to mitigate the effects of climate change grow, reliable and thorough reporting of greenhouse gas emissions are essential for measuring progress towards international and domestic emissions reductions targets. New Zealand's national emissions inventories are currently reported between 15 to 27 months out-of-date. We present a machine learning approach to nowcast (dynamically estimate) national greenhouse gas emissions in New Zealand in advance of the national emissions inventory's release, with just a two month latency due to current data availability. Key findings include an estimated 0.2\% decrease in national gross emissions since 2020 (as at July 2022). Our study highlights the predictive power of a dynamic view of emissions intensive activities. This methodology is a proof of concept that a machine learning approach can make sub-annual estimates of national greenhouse gas emissions by sector with a relatively low error that could be of value for policy makers.
\end{abstract}
    
\maketitle

\section{Introduction}\label{sec:introduction}

Emissions from human activity are the key driver of the anthropogenic climate change observed since the onset of the industrial revolution (c. 1850). In an effort to mitigate our influence over the global climate, the Paris Climate Agreement aims to limit global temperature rise to well below 2°C and pursue efforts to limit warming to 1.5°C above pre-industrial levels, requiring significant action to reduce greenhouse gas emissions to net zero on an international scale. Reliable and thorough greenhouse gas inventories are fundamental in measuring progress towards these targets and provide insight into domestic and international emissions reduction policies and strategies~\cite{yona2020refining}. Guidelines for such inventories are published by the Intergovernmental Panel on Climate Change (IPCC) and were refined in 2019, the first major update since 2006~\cite{Jamsranjav17}. We develop a proof of concept that combines machine learning methods with indicators of dominant trends in New Zealand's emissions profile in order to improve timeliness of emissions reporting with relatively low error, and therefore `nowcast' New Zealand's emissions in the sense described in~\cite{nordic_nowcasting}. The importance of greater timeliness is also highlighted in~\cite{yona2020refining}.

New Zealand's national emissions are reported in the New Zealand Greenhouse Gas Inventory (NZGGI). It is the official annual report of all anthropogenic emissions and removals of greenhouse gases (GHGs) associated with production, which is currently released approximately 15 months after the end of the calendar year being reported on, as determined by international reporting guidelines~\cite{MfE_national_inventory_report_2020}. Multiple GHG types estimated by the NZGGI are aggregated in terms of CO\textsubscript{2} equivalents (CO\textsubscript{2}-e).\footnote{Various research debates ways of aggregating GHGs (see~\cite{RS21} and references therein). We are bound to CO\textsubscript{2}-e since it is used by the NZGGI.} Our indicator-based model estimates CO\textsubscript{2}-e directly rather than modelling each GHG type to compensate for limited data availability.

National emissions inventories are time-consuming to produce. At any given time, national GHG emissions reported in the NZGGI are between 15 and 27 months out of date. Methodologies for the release of quarterly and more timely emissions estimates are in the process of being developed in a number of countries \cite{huang2021highly,nordic_nowcasting,SN21}. Within New Zealand, Statistics New Zealand are developing quarterly emissions reporting~\cite{SNZ_quarterly_emissions_example} but, as we discuss in \cref{sec:local quarterly emissions reports}, this is not comparable to the NZGGI in many ways. This existing gap highlights the need for a more dynamic dataset that complements the NZGGI to be developed.

Numerous studies exist that estimate daily emissions (i.e. emissions occurring within any given day) by first developing a daily inventory of emissions. \cite{forster2020current} develops a `bottom-up' inventory using national mobility data, finding that emissions reduced during the COVID-19 pandemic, by as much as 30\% during April 2020. \cite{liu2020near} constructs a measure of CO\textsubscript{2} emissions by developing an inventory by country comprised of the sectors that are power generation, industry, transportation and household consumption, each broken down by the three fossil fuel types which are coal, oil and natural gas. Findings include that CO\textsubscript{2} emissions decreased by 8.8\% in the first half of 2020 compared to the same period of the previous year in China. \cite{le2020temporary} again develops an inventory using a `bottom-up' approach, comprised of six sectors, finding that daily CO\textsubscript{2} emissions decreased by -17\% by April 2020 compared to the mean 2019 levels, with surface transport being the sector that influenced reduction most heavily. Further studies that estimate daily emissions by developing a daily inventory of emissions include~\cite{liu2020carbon,oda2021errors,zheng2020satellite}. Such inventories may be labour-intensive to curate, even more so than the NZGGI at the annual level. These studies have impressively specific results, such as daily emissions estimates with a regional component while separating CO\textsubscript{2} and air pollutants~\cite{huang2021highly}. However, they need to focus on short windows of time to obtain a sufficiently detailed inventory in the first place. This distinguishes our work, which reports on emissions from 1990 up to two months prior to the present day.

We produce daily estimates of cumulative GHG emissions in every 365 day period since 1990 with a latency of less than two months. In other words, we produce daily estimates of annual emissions (where annual is thought of as a rolling 365 day window). Since we provide daily estimates of annual emissions, the estimate occurring on the last day of a given year is an estimate for the cumulative emissions that year. This figure is comparable to the annual value coming from the NZGGI for that year.

We compare a number of potential models in terms of prediction accuracy, model stability and over-fitting.
The approach can be applied to any sector and has so far been applied to the Energy and Agriculture sectors of the NZGGI~\cite{MfE_national_inventory_report_2020}. These sectors were chosen to begin with for their relative contribution to the NZGGI, making up 90.0\% of emissions in 2020~\cite{MfE_emissions_tracker}.

The input variables for each sector are called \textit{indicators}. There is a growing precedent that a relatively small number of indicators can be sufficient to model broad greenhouse gas emissions systems, recent examples including~\cite{IAS22,Ulk22}. Using this approach, we are able to model the Energy and Agriculture sectors with just four and six indicators, respectively. Our models do not assume a relationship between indicators and the emissions of their sectors. Thus, to explain the inner workings of the model we have to infer relationships between indicators and predictions by the model. We use subject knowledge to verify that these relationships are sensible. In addition, we ensure our methodology is repeatable between editions of the NZGGI.

We assess uncertainties carefully in terms of accuracy, stability and over-fitting in Sections~\ref{sec:accuracy} to \ref{sec:overfitting}. Additionally, we compare our uncertainties to typical inter-annual variation in \cref{sec:interannual} to mitigate the concern that the error due to our model exceeds expected changes from one year to the next. Results are derived in \cref{sec:results} using both of our high frequency and low latency of emissions reporting. The insights resulting from our models could improve current understanding of emissions in a landscape changing day to day.

\section{Methods}

\subsection{Overview}\label{sec:methodology_overview}

We use cross validation to compare a number of potential models that predict NZGGI sector emissions using data on variables related to sector emissions. A model selection protocol based on prediction accuracy, model stability and over-fitting is used to decide a preferred model. We apply the preferred model to estimate the Energy and Agriculture sectors of the NZGGI~\cite{MfE_national_inventory_report_2020}. These sectors were chosen to begin with for their size, together making up 90.0\% of emissions in 2020~\cite{MfE_emissions_tracker}. We call the independent variables used for each sector \textit{indicators}, which we describe in detail in \cref{sec:indicators}. Our models are non-parametric in the sense that we do not assume a data model. Thus we face the risk of selecting a model that cannot be explained. We combat this risk in \cref{sec:explanation}.

The main objective of the model is to attain high frequency and low latency outputs. Because of limited data availability, to achieve the objective we make some simplifying choices: (1) as described above, our dependent variables are sector emissions (rather than estimating every source of emissions and then aggregating) and (2) we estimate carbon dioxide equivalents directly (rather than estimating each GHG type individually and then aggregating such as in~\cite{huang2021highly,Ulk22}).

The NZGGI updates annually. The latest edition at the time of writing reports until 2020, however our analysis is applicable to all editions of the NZGGI. Therefore, we discuss the model mostly with respect to any edition of the NZGGI. Any results presented pertain to the 2020 edition~\cite{MfE_spreadsheet}.

Existing research involves data sets of a wide variety of frequencies and time-spans. The most similar studies to ours in these terms are~\cite{JPL22,Ulk22,WEN2020137194}, which have an annual frequency and time-spans of 1990-2018, 1990-2019 and 1997-2017, respectively. We have a time-span of 1990-2020 when using the latest edition of the NZGGI. We collect new indicator data daily so that predictions using the model can be made daily. This means our outputs not only estimate year-end values that mimic the NZGGI but also interpolate between those year-end estimates and leave us with a low latency of less than two months in practice. A more detailed description of our daily estimates of emissions is given in \cref{sec:introduction}.

\subsection{Indicators}\label{sec:indicators}

\textit{Indicators} are times series associated with natural variables that are chosen as input variables for the model of a given sector. There are many variables that could serve as indicators of emissions in our model. The indicators discussed in this paper are non-exhaustive and this model could in fact be replicated with an entirely different set. We therefore do not claim they are the most important of the potential indicators, but they are sufficient in the sense that they effectively model their sectors at this point in time and are available dynamically. Our indicators are chosen for: (1) their systematic relevance to GHGs, (2) a resulting relationship with sector emissions predictions that is sensible, and (3) the availability of the necessary data. The indicators used in the model are summarised in \cref{tab:indicators}.

\begin{table}
    \tiny

    \begin{tabularx}{\textwidth}{L{0.5}L{1}L{1}L{1}L{1}L{1}L{1}} 
    
    Code & Name & Unit & Frequency & Sector & Direct/indirect & Data sources \\
    
    \midrule
    
    LEE & \LEE & [1] & Daily & Energy & Direct &~\cite{em621,NZTA21b} \\

    REM & \REM & [kgCO\textsubscript{2}-e/km] & Monthly & Energy & Direct &~\cite{MfE_detailed_guide,NZTA21a} \\

    ETV & \ETV & [kgCO\textsubscript{2}-e] & Daily & Energy & Direct &~\cite{scrappage,MfE_detailed_guide,NZTA21a,NZTA21b} \\

    COA & \COA & [tonnes] & Quarterly & Energy & Indirect &~\cite{MBIE21} \\

    COW & \COW & [number] & Annual & Agriculture & Indirect &~\cite{AGR001AA} \\

    MEA & \MEA & [tonnes] & Weekly & Agriculture & Indirect &~\cite{NZMB21,LSS025AA} \\

    FIM & \FIM & [tonnes] & Monthly & Agriculture & Indirect &~\cite{IMP033AA} \\

    EXM & \EXM & [\$] & Monthly & Agriculture & Indirect &~\cite{EXP012AA} \\

    EXD & \EXD & [\$] & Monthly & Agriculture & Indirect &~\cite{EXP012AA} \\

    EXF & \EXF & [\$] & Monthly & Agriculture & Indirect &~\cite{EXP012AA} \\
    
    \end{tabularx}

    \caption{Indicators used to model sectors. Direct means the indicator considers the emissions intensity of the underlying activity. Definitions of the indicators are given in \cref{appendix:indicators}.}
    \label{tab:indicators}

\end{table}

Each indicator represents an emissions intensive activity (e.g. kgCO\textsubscript{2}-e per km of petrol-fueled vehicle travel). Intensity (i.e. emissions per unit of activity) helps to account for the necessary dissociation of emissions with activity and productivity. If an indicator considers intensity, then we say it is \textit{direct}. 
For example, one of our indicators measures the emissions potentials of new vehicles on New Zealand roads based on motive power, year of manufacture and gross vehicle mass. The data comes from a national vehicle register that is updated monthly. Therefore, our model accounts for variation in emissions intensity over time via indicators such as these. For more information, see \cref{sec:newly_registered_emissions}. If an indicator is not direct, we say it is \textit{indirect}. We use a combination of direct and indirect indicators.

We check for updates to our indicator data on a daily basis to ensure the data is as up to date as possible. Where a dataset is not updated on a daily basis, the dataset is held constant until the present day and interpolated linearly between historic data points in order to provide a daily value. In the extreme case where a dataset is updated only annually, this would mean the data is held constant for up to 365 days.

In \cref{appendix:indicators}, we describe the construction of each indicator in detail. Given the hundreds of sources of emissions featured in the NZGGI, the indicators presented in \cref{tab:indicators} appear scarce. We acknowledge there are many potential indicators, and as new datasets emerge, these may model the sectors better than those datasets currently being used. While the indicators used here are not necessarily optimal, they are sufficient for this proof of concept. It is important to note, as previously discussed, that the benefit of this approach is the limited number of variables required within the model. 

\begin{figure}[H]
    \centering
    \begin{subfigure}[t]{0.45\textwidth}
        \includegraphics[width=\textwidth]{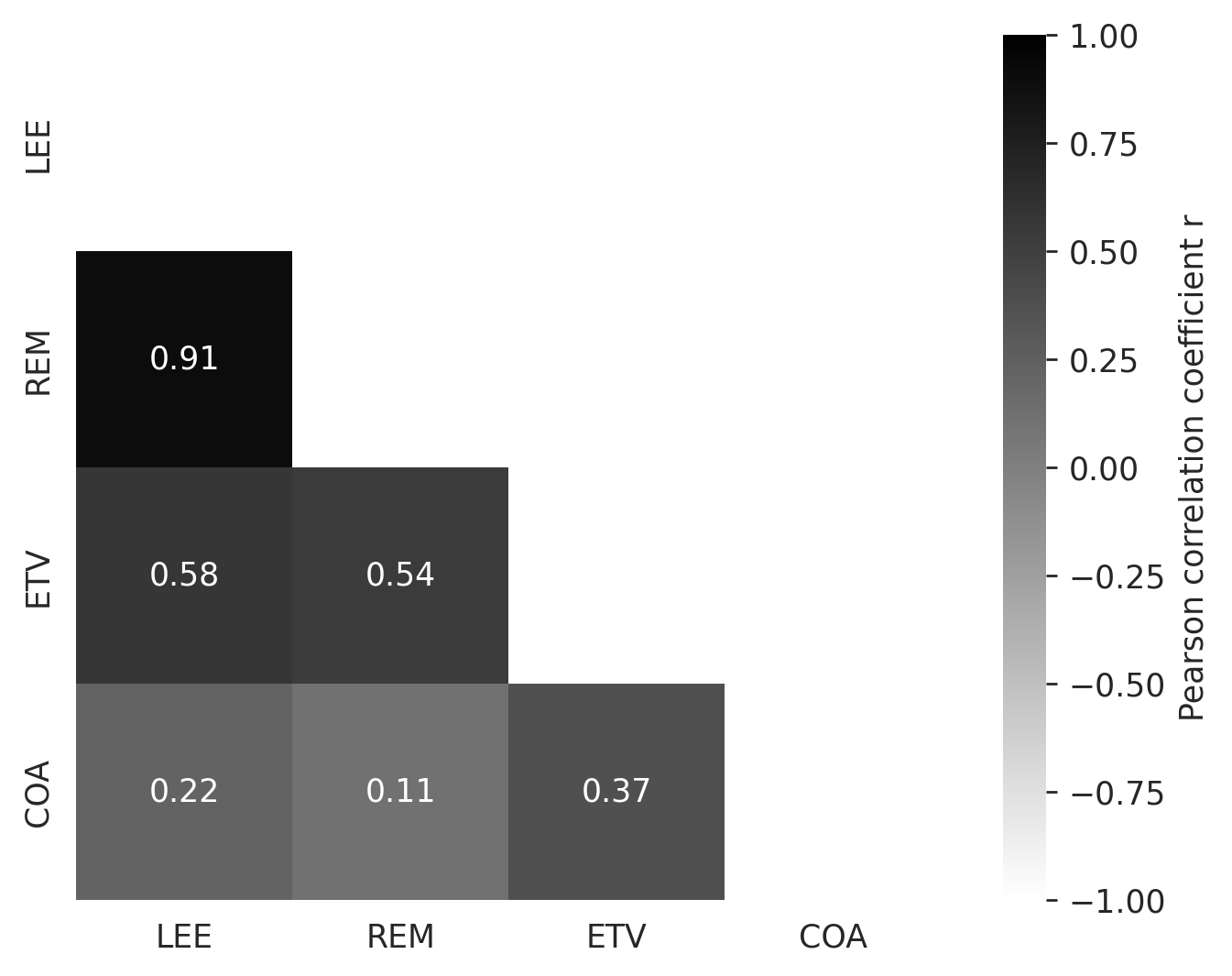}
        \caption{Energy. LEE: \LEE, REM: \REM, ETV: \ETV, COA: \COA.}
        \label{fig:corr_Energy}
    \end{subfigure}
    \hspace{1cm}
    \begin{subfigure}[t]{0.45\textwidth}
        \includegraphics[width=\textwidth]{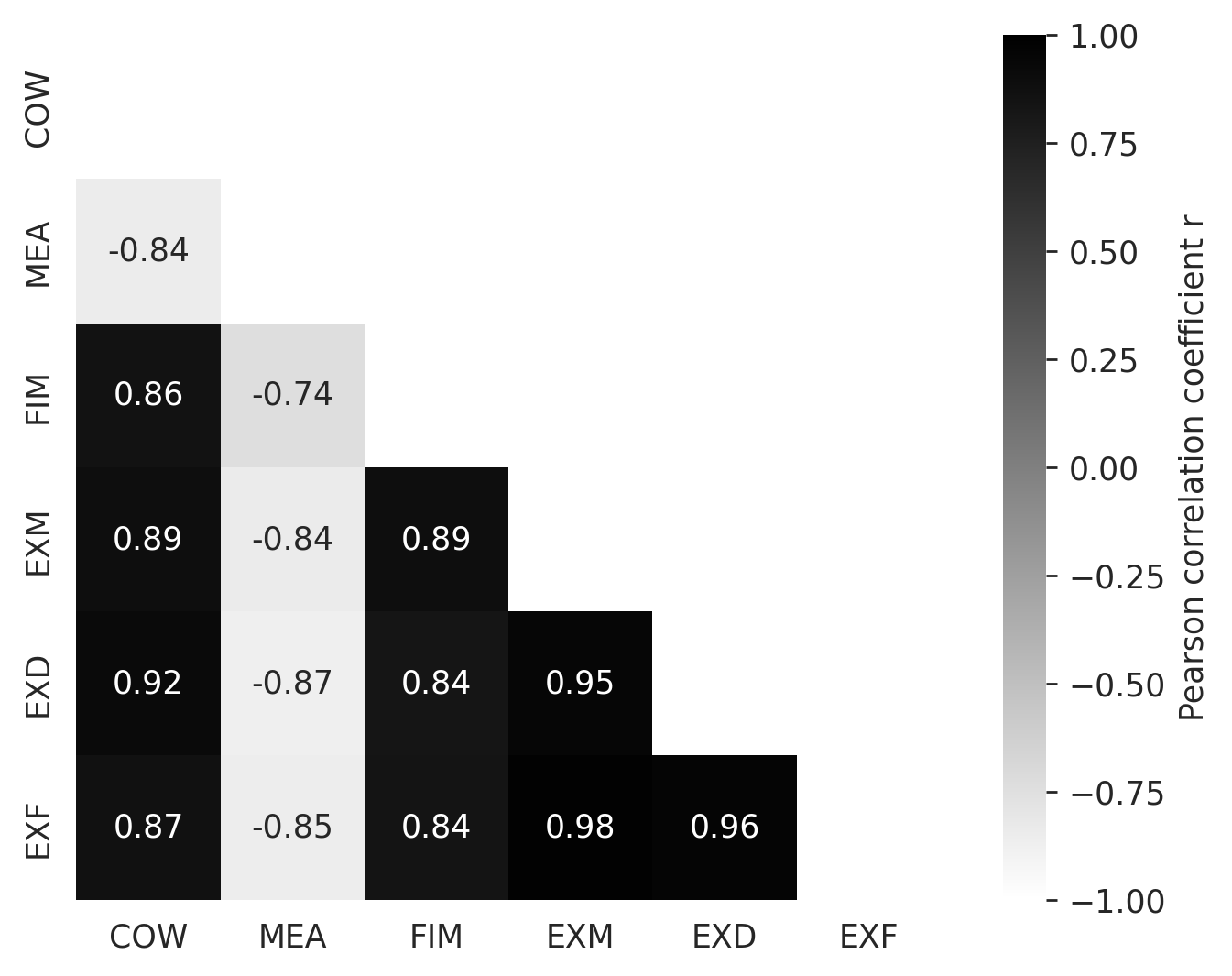}
        \caption{Agriculture. COW: \COW, MEA: \MEA, FIM: \FIM, EXM: \EXM, EXD: \EXD, EXF: \EXF.}
        \label{fig:corr_Agriculture}
    \end{subfigure}
    \caption{Heatmap of correlation matrix showing high collinearity between indicators in each sector.}
    \label{fig:sub_figures_1}
\end{figure}

Furthermore, the performance of regression models is generally sensitive to collinearity between indicators~\cite{farrar1967multicollinearity}. The scarcity of useful data makes it difficult to find indicators with low collinearity. We note that there exists high collinearity between indicators as shown in \cref{fig:sub_figures_1}. While methods exist to aggregate correlated variables, we keep our indicators as individual inputs for simplicity. Therefore, we will prefer machine learning models that have no such requirements around collinearity of indicators. Certain machine learning models also allow non-linear relationship between indicators and sectors~\cite{Ulk22}. This is valuable because including non-linear indicators can be critical for prediction accuracy~\cite{WEN2020137194}.

We note that it is not intuitive that there is a negative correlation between MEA and EXM. However, there are possible explanations for this relationship, such as the export-inspected facilities where MEA is counted having a limited scope or the variation in the value of exports over time outweighing an actual decrease in quantity of meat exports. This requires further investigation. We discuss this further in \cref{sec:explanation}.

\subsection{Model selection}\label{sec:sector_models}

\subsubsection{Models}\label{sec:model_types}

We compare three models in our selection process: bounded-variable least squares~\cite{SP95} (LS), random forests \cite{Bre01} (RF) and extra trees~\cite{GEW06} (ET).

Bounded-variable least squares~\cite{SP95} (LS) solves the usual linear least squares problem with the added constraint that fitted coefficients lie within user-specified bounds. We use the Python programming language~\cite{Python3} for our analysis. In particular, LS is implemented with a linear least squares solver from the SciPy project~\cite{SciPy}. For each indicator, we set the bounds to either $[0,\infty)$ or $(-\infty,0]$ depending on the known systematic relationship between indicator and emissions. This aligns with the use of a `relation expected sign' as in~\cite[Table 3]{GSMO20}. \cite{IAS22} use such bounds to classify indicators as `enablers' or `de-enablers'. Significant statistical analysis is required to verify the use of LS, as demonstrated in~\cite{GSMO20}. As mentioned in \cref{sec:indicators}, collinearity and non-linearity of indicators makes this infeasible in our setting. In selecting our optimal model, we will select against LS anyway because it is outperformed by the other models in terms of prediction accuracy (see \cref{tab:test_mape}). This suggests that the latent relationships are complex and/or non-linear. We therefore omit statistical details relevant only to LS.

Random forest~\cite{Bre01} (RF) is an ensemble method that builds a collection of decision trees by incorporating a random component into their construction and makes a prediction by aggregating the predictions of each tree. The technique is \textit{non-parametric} in the sense that RF does not assume a data model. However, there are various hyper-parameters for the structure of the forest and its trees. These are manipulated in order to \textit{tune} the model to optimise its predictions. This is discussed in detail in \cref{sec:cross_validation}. We implement RF using the random forest regressor by Scikit-Learn~\cite{sklearn}.

We elaborate on the random component mentioned above in the construction of each tree to aid the description of the next model. Each decision tree in the forest involves a random sample of indicators and data points. In RF, these are sampled with replacement. In our application we have only 31 data points to begin with (each year from 1990 to 2020). Hence, we expect we cannot afford sampling with replacement in the random component described above. This motivates us to use ET. 

Extra trees~\cite{GEW06} (ET) is similar to RF but is different as follows. While the trees in RF involve sampling with replacement, the trees in ET involve sampling without replacement. Thus each tree in ET is likely to learn from a more comprehensive sample of data. In a similar study to ours~\cite{JPL22}, RF did not perform exceptionally well, placing fourth among eight models considered. We show in \cref{sec:accuracy} that our implementation of RF is more accurate than~\cite{JPL22}. We observe in \cref{sec:stability} that stability issues of RF are resolved by our inclusion of ET. We use the extra trees regressor by~\cite{sklearn} to implement ET.

\subsubsection{Cross validation}\label{sec:cross_validation}

\textit{Over-fitting} occurs when a fitted model performs worse on unseen data than on its training data. Cross validation is a technique used to assess for over-fitting. We follow a standard workflow that incorporates hyper-parameter tuning when applicable.

Since the tree-based methods have hyper-parameters to be tuned, cross validation occurs at two levels: (1) validating to find the best hyper-parameters and (2) testing to ensure the model is not over-fitting after retraining on the best hyper-parameters. Thus we use a workflow that is best described as `repeated double cross validation' by~\cite{FLV09} (see also~\cite[Figure 7.4]{Huang2022}). This means that at the lowest level the data is split into three parts: pre-training, validating and testing. We use the train-test split utility by~\cite{sklearn} with a test size of 6 out of the 31 data points. This gives a train:test ratio of about 80:20, which is the same ratio as in~\cite{WEN2020137194}. If the model on hand is RF or ET, then hyper-parameters need tuning. Randomised search through a given hyper-parameter space is used to find the best hyper-parameters. We use randomised search instead of its predecessor, grid search, because randomised search is known to be better at navigating situations where hyper-parameters have differing importance~\cite[Figure 1]{BB12}. Randomised search cross validation by~\cite{sklearn} with $5$ folds implements the randomised search, which repeatedly splits the 25 training data points into 20 for pre-training and 5 for validating.

The hyper-parameter space used both for RF and ET is defined with the ranges:
\begin{itemize}
    \item Number of trees: $\{200,201,...,219\}$,
    \item Maximum depth of tree: $\{1,2,3\}$,
    \item Minimum number of samples required to split: $\{2,3,4\}$, and
    \item Minimum number of samples required to be at a leaf node: $\{2,3,4\}$.
\end{itemize}
These were chosen by running a large number of investigations to find a large hyper-parameter space in which the models were stable in the sense we describe in \cref{sec:model_selection}.

Fitting each model on a particular train-test split of the data (as described above) defines a single simulation in our cross validation. We run 30 such simulations. This makes our cross validation more comprehensive than \cite{JPL22,Ulk22}, which consider only one simulation each.

\subsubsection{Model outcomes}\label{sec:model_selection}

The objective is to select the optimal model in terms of certain `model outcomes'. The model outcomes used in~\cite{CMW18} are: over-fitting, prediction accuracy and execution time. The bounds on the parameters, described earlier, ensure that the use of the attributes remains aligned with any latent data structures. Execution time is not important to us since our data sets are small, so we omit this outcome.

A novelty in our approach is to include an outcome about the stability of the structure of the model. Different models have different structures, so the notion of stability has to adapt to each model. For LS, we consider the distribution during cross validation of the fitted coefficients in the linear model. For RF and ET, instead of coefficients, we consider the values of indicator importance in the tuned estimator. A model is considered more stable if it has a smaller variation in the distribution above. These are described in more detail in \cref{sec:stability} and pictured in \cref{fig:feature_importance}.

Thus, our objective is to find the optimal model in terms of the model outcomes:
\begin{enumerate}
    \item Prediction accuracy, in terms of mean absolute percentage error (MAPE):
    \[
        \frac{100}{n} \times \sum_{x\in X_{\text{test}}}\left|\frac{x - \hat{x}}{x}\right|
    \]
    where $\hat{x}$ is the prediction for each $x$ in a given test set $X_{\text{test}}$ of size $n$ in cross validation, like in~\cite{CMW18},
    \item Model stability, in terms of distribution of fitted coefficients/indicator importance as described above, and
    \item Over-fitting, in terms of the average difference of MAPE between training and testing data, like~\cite{CMW18}.
\end{enumerate}

\section{Results}\label{sec:results}

\subsection{Framework to assess models to predict sector-level greenhouse gas emissions in national inventories}\label{sec:framework}

Cross validation was applied to compare a range of models of greenhouse gas emissions by sector in terms of a comprehensive list of model outcomes (\cref{sec:sector_models}): prediction accuracy, stability and over-fitting. A machine learning model, extra trees (\cref{sec:model_types}), was found to perform the best according to our selection protocol (\cref{sec:final_selection}). Relationships between sector emissions and indicators were inferred from the model (\cref{sec:explanation}), and subject knowledge was used to explain these relationships. Outliers in the relationships were identified via Mahalanobis distance and explained in context.

\subsubsection{Prediction accuracy}\label{sec:accuracy}

Statistics for prediction accuracy are given in \cref{tab:test_mape}, which shows ET is optimal across both sectors.

{\scriptsize\begin{table}[H]
\centering

\begin{tabular}{p{1cm}p{1cm}p{1cm}p{1cm}p{1cm}p{1cm}p{1cm}}
\toprule
sectors & \multicolumn{3}{l}{Agriculture} & \multicolumn{3}{l}{Energy} \\
model &          LS &        RF &        ET &        LS &        RF &        ET \\
statistics &             &           &           &           &           &           \\
\midrule
count      &    30 &  30 &  30 &  30 &  30 &  30 \\
mean       &       1.962 &     1.484 &     \textbf{1.347} &     3.818 &     3.565 &     \textbf{3.211} \\
std        &       0.560 &     0.634 &     0.622 &     0.837 &     0.911 &     0.866 \\
min        &      \textbf{0.444} &     0.448 &     0.526 &     2.451 &     1.732 &     \textbf{1.602} \\
25\%        &       1.594 &     1.162 &     \textbf{1.071} &     3.356 &     3.167 &     \textbf{2.732} \\
50\%        &       1.999 &     1.406 &     \textbf{1.314} &     3.664 &     3.617 &     \textbf{3.194} \\
75\%        &       2.244 &     1.635 &     \textbf{1.455} &     4.452 &     4.023 &     \textbf{3.779} \\
max        &       \textbf{3.199} &     3.844 &     3.967 &     5.612 &     6.232 &     \textbf{5.583} \\
\bottomrule
\end{tabular}
\caption{Descriptive statistics for MAPE (\%) values in test set for each of the 30 simulations. LS: bounded-variable least squares, RF: random forest, ET: extra trees. A value for a summary statistic (e.g. mean) is in bold if it is optimal compared to other models in the same sector. This gives an indication of which model is optimal while accounting for the distribution of MAPE values observed throughout simulation.}
\label{tab:test_mape}
\end{table}
}

Our results are comparable to those found in \cite{JPL22} due to the similarity in methods and initial data. In both settings, MAPE is used to measure error when predicting annual values of CO\textsubscript{2}-e. Our models have resulted in higher accuracy than those presented \cite{JPL22}. The comparable model to LS in~\cite{JPL22} is `ElasticNet' since it too is a regularised linear model. The comparable model to both RF and ET in \cite{JPL22} is `Random Forest'. LS achieves a median MAPE of $2.0\%$ in Agriculture and $3.7\%$ in Energy, whereas ElasticNet achieves a MAPE of $19.462\%$ in one application of the model and $41.619\%$ in another \cite[Tables II and III]{JPL22}. ET (respectively, RF) achieves a median MAPE of $1.3\%$ (respectively, $1.4\%$) in Agriculture and $3.2\%$ (respectively, $3.6\%$) in Energy, whereas Random Forest in~\cite{JPL22} achieves a MAPE of $9.470\%$ in one application and $21.522\%$ in another \cite[Tables II and III]{JPL22}. Our higher accuracy is possibly due to our use of hyper-parameter tuning. There is no discussion of tuning in \cite{JPL22}.

\subsubsection{Model Stability}\label{sec:stability}

\begin{figure}[H]
    \centering
    \begin{subfigure}[t]{0.45\textwidth}
        \includegraphics[width=\textwidth]{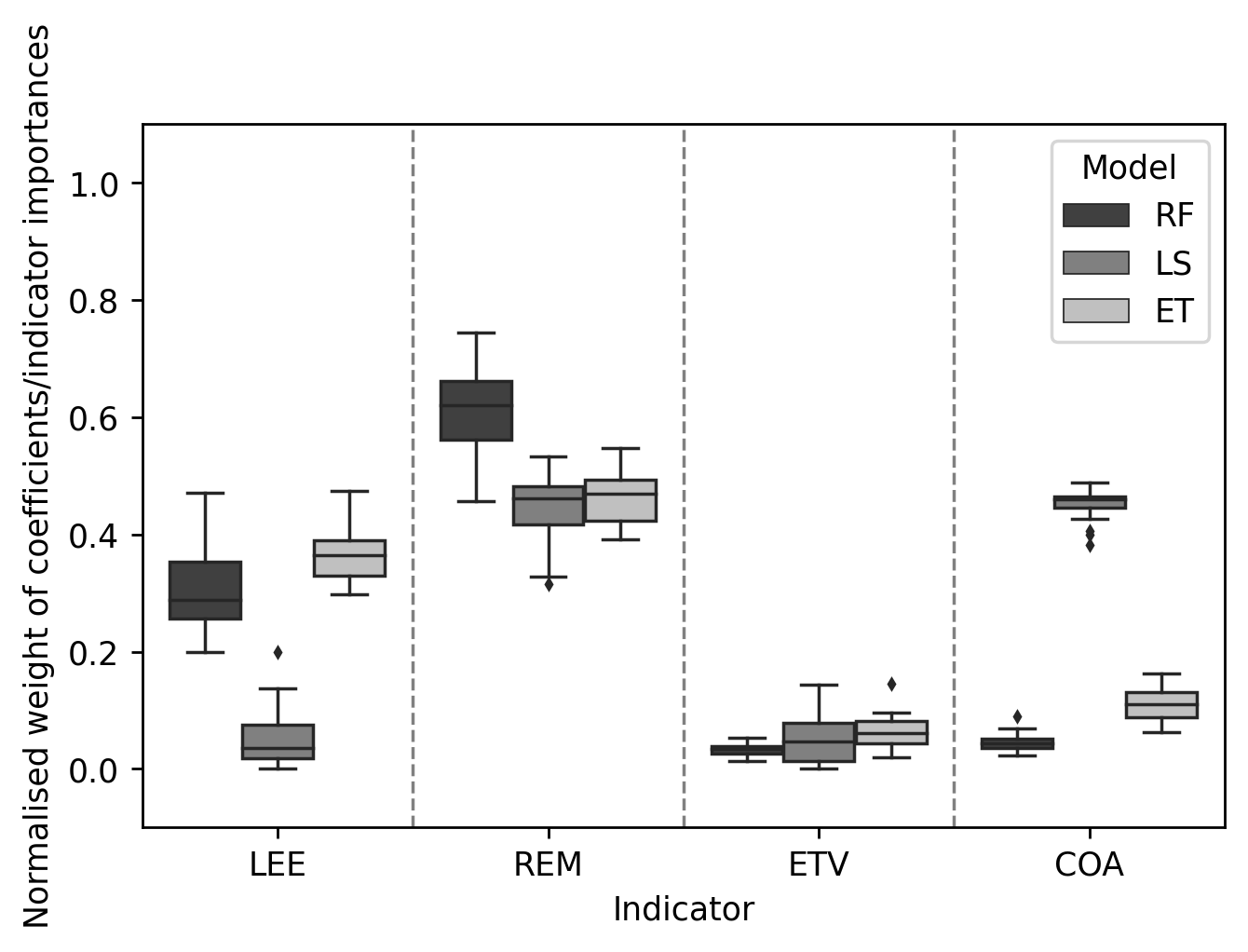}
        \caption{Energy: LEE: \LEE, REM: \REM, ETV: \ETV, COA: \COA.}
        \label{fig:indicator_importance_en}
    \end{subfigure}
    \hspace{1cm}
    \begin{subfigure}[t]{0.45\textwidth}
        \includegraphics[width=\textwidth]{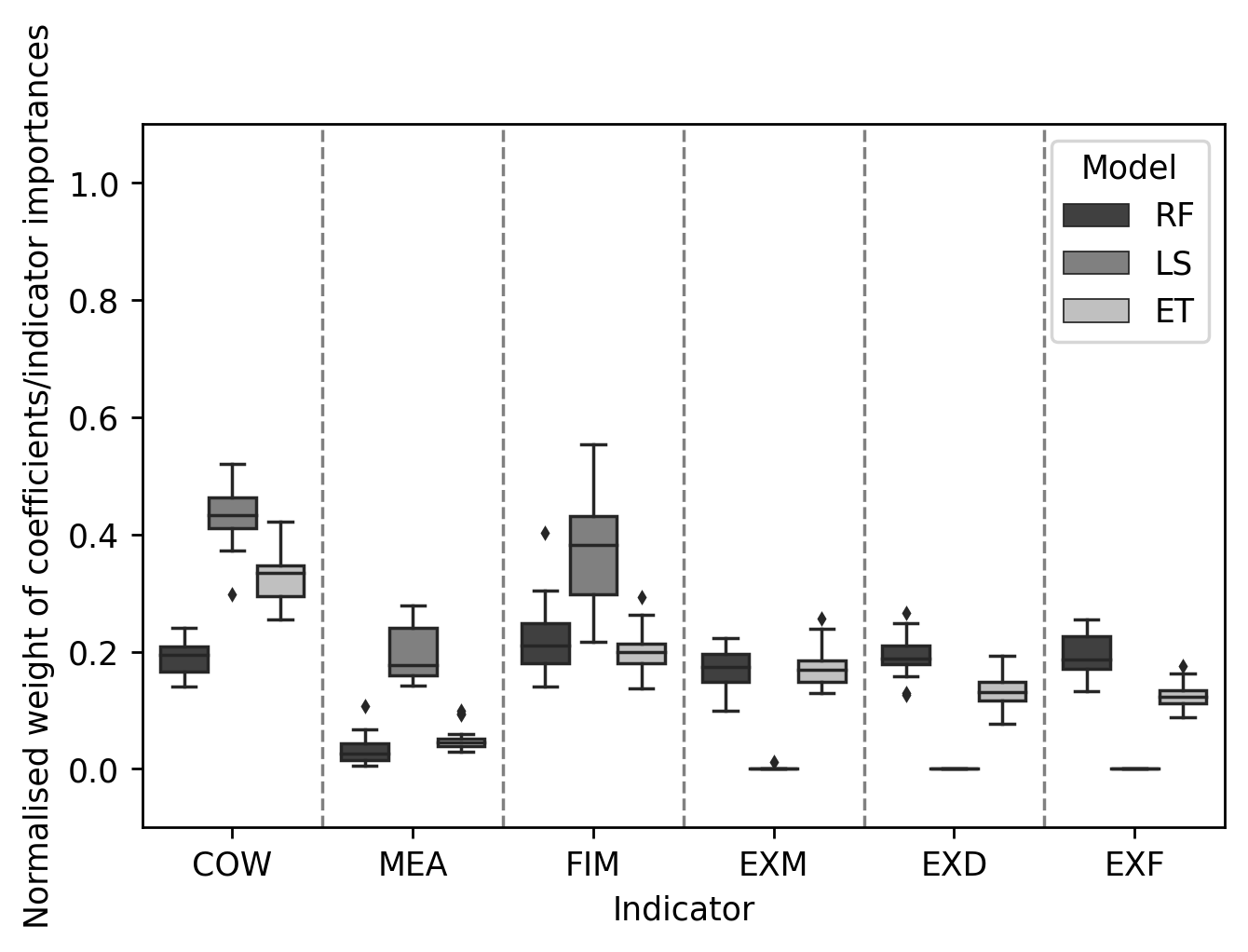}
        \caption{Agriculture: COW: \COW, MEA: \MEA, FIM: \FIM, EXM: \EXM, EXD: \EXD, EXF: \EXF.}
        \label{fig:indicator_importance_ag}
    \end{subfigure}
    \caption{Box plots picturing distribution of fitted coefficients (for LS, bounded-variable least squares) and indicator importance (for RF and ET, random forest and extra trees, respectively) values as a measure of stability of each model.}
    \label{fig:feature_importance}
\end{figure}

Recall from \cref{sec:model_selection} that LS (respectively, ET and RF) is considered to be more stable if it has smaller variation in the distributions of its fitted coefficients (respectively, indicator importance values). These distributions are pictured in \cref{fig:feature_importance}. We observe stability can vary in a number of ways.

Firstly, the order of indicators by median value depends on the model. For example, in RF, the median value for FIM (\FIM) is slightly higher than that of COW (\COW), whereas in ET the box and whiskers for COW is entirely above the upper quartile for FIM.

Secondly, notice simply how long the boxes and whiskers in some models are, and therefore how unstable the models can be. ET has consistently short boxes and whiskers. On the other hand, in RF for Energy, the length of the box and whiskers for the two most important indicators are long, especially in comparison to the box and whiskers for the same indicators under ET. This instability in RF is undesirable since it demonstrates high uncertainty about the impact of each indicator in the model.

Lastly, models can be more or less `polarising' about the coefficient/importance of an indicator. Notice LS is very polarising since it tends to assign either very high or low coefficients. This is particularly evident in the Agriculture sector (see \cref{fig:indicator_importance_ag}); three of the six indicators' coefficients are always $0$. This annihilation of indicators is a common feature of a LS because of the bounding of variables. This is likely to be a manifestation of collinearity, highlighting the importance of variable stability as one of our model outcomes.

In summary, we observe in \cref{fig:feature_importance} that ET is the most stable compared to RF and LS.

\subsubsection{Over-fitting}\label{sec:overfitting}

LS had the least over-fitting compared to ET and RF. That is, the maximum difference in MAPE between training and testing in LS was $1.9\%$ compared to $3.1\%$ for both ET and RF in Agriculture and was $2.8\%$ compared to $4.2\%$ for RF and $3.6\%$ for ET in Energy (see \cref{tab:test_and_train_mape_difference}). That said, all models performed well as demonstrated by means in \cref{tab:test_mape}. These values are acceptable in that similar or higher values of over-fitting are expressed in~\cite[Table 4]{CMW18}.

{\scriptsize\begin{table}
\centering
\begin{tabular}{p{1cm}p{1cm}p{1cm}p{1cm}p{1cm}p{1cm}p{1cm}}
\toprule
sectors & \multicolumn{3}{l}{Agriculture} & \multicolumn{3}{l}{Energy} \\
model &          LS &        RF &        ET &        LS &        RF &        ET \\
statistics &             &           &           &           &           &           \\
\midrule
count      &    30 &  30 &  30 &  30 &  30 &  30 \\
mean       &       0.517 &     0.696 &     0.561 &     0.804 &     1.743 &     1.264 \\
std        &       0.677 &     0.662 &     0.620 &     1.006 &     0.994 &     0.999 \\
min        &      -1.339 &    -0.336 &    -0.324 &    -0.835 &    -0.232 &    -0.594 \\
25\%        &       0.093 &     0.344 &     0.255 &     0.239 &     1.379 &     0.748 \\
50\%        &       0.584 &     0.603 &     0.537 &     0.617 &     1.873 &     1.223 \\
75\%        &       0.938 &     0.894 &     0.714 &     1.605 &     2.318 &     1.881 \\
max        &       1.865 &     3.126 &     3.069 &     2.834 &     4.157 &     3.639 \\
\bottomrule
\end{tabular}
\caption{Descriptive statistics for MAPE difference (\%) between test set and train set for each simulation. LS: bounded-variable least squares, RF: random forest, ET: extra trees.}
\label{tab:test_and_train_mape_difference}
\end{table}
}

Recall from \cref{sec:cross_validation} that there are only 31 objects in our dataset. It is well known that over-fitting is more common for smaller samples. Yet we successfully mitigated over-fitting as above. Moreover, aside from~\cite{CMW18}, we found it is not common in the literature to explicitly analyse over-fitting at all. For example,~\cite{Mut22,MKSK22,JPL22,Ulk22} do not mention over-fitting. The study~\cite{WEN2020137194} mentions over-fitting and the use of training and testing sets but does not report any measures of over-fitting.

\subsubsection{Selection}\label{sec:final_selection}

Our selection protocol is as follows: given the models LS, RF and ET, remove the model performing worst in terms of prediction accuracy; from the remaining two models remove the model performing worst in terms of model stability; the remaining model is selected provided it is not over-fitting.

Over-fitting is regarded as a lower priority compared to accuracy because prediction accuracy is measured on the test set, so it mitigates over-fitting implicitly. Also, over-fitting is regarded as a lower priority than stability since an unstable model cannot be explained. In \cref{sec:comparison_to_other_methods}, we compare our selection protocol to that in~\cite{CMW18}.

We apply the selection protocol now. The measures in \cref{tab:test_mape} demonstrate LS is the least accurate, so we remove LS from the running. The accumulated length of RF box and whiskers (\cref{fig:feature_importance}) is greater than that of ET, so RF is removed for being less stable than ET. The remaining model is ET and we recall that its over-fitting values were acceptable.

\subsection{Uncertainty in model compared to inter-annual variation}\label{sec:interannual}

Throughout \cref{sec:framework} we recorded uncertainties of our models from the point of view of the model outcomes established in \cref{sec:model_selection}. In particular, this involved optimising prediction accuracy by minimising residuals. However, no matter how small the residuals are, an outstanding concern is whether the residuals exceed typical levels of inter-annual variation of NZGGI emissions estimates, in which case the predictions would be useless since the accuracy of the predictions would be outweighed by expected variation between years.

We have mitigated this concern by comparing our residuals to inter-annual variation (measured in terms of year-on-year differences in NZGGI emissions estimates). For both sectors, the maximum, upper quartile, median and lower quartile values are all significantly lower for our residuals than for inter-annual variation. The following values are given as percentages of the average value for the sector. For the Energy sector, the lower quartile, median and upper quartile for our residuals are 0.54\%, 1.73\% and 2.89\%, respectively, whereas they are 1.20\%, 4.08\% and 5.04\% for inter-annual variation. For the Agriculture sector, the lower quartile, median and upper quartile for our residuals are 0.26\%, 0.54\% and 1.27\%, respectively whereas they are 0.47\%, 1.00\% and 1.71\% for inter-annual variation. That is, we find the uncertainty of the model tends to be significantly below typical inter-annual variation.

We also consider the following year-wise paired view comparing inter-annual variation to our residuals. For each year, we divide the relevant value of inter-annual variation with our residual that year, and compute the median of these fractions. The median for the Energy sector was 1.804 (meaning inter-annual variation tends to be 80\% higher than our residuals), while the median for the Agriculture sector was 1.322 (meaning inter-annual variation tends to be 32\% higher than our residuals).

\subsection{Explanation of model}\label{sec:explanation}

Our selected model, ET, does not presume a data model. Hence, in order to check the model is sensible/can be explained, we must infer relationships between inputs and outputs a posteriori. A relationship is \textit{inferred} between an indicator and its sector if a multivariate trend is present. We find every indicator has an inferred relationship with its sector. Each inferred relationship is tested for outliers by calculating Mahalanobis distances~\cite{Ghorbani2019} and calculating associated $p$-values on a $\chi^2$-distribution with $1$ degree of freedom. Then, inferred relationships and their outliers are explained using subject knowledge.

Inferred relationships must be understood in the context of the indicators used. Changing the selection of indicators would lead to variation in the inferred relationships. In this way, our methodology has a limited capacity to make precise/absolute claims about drivers of emissions. With this in mind, we give two examples of inferred relationships and their explanations.

\begin{figure}
    \centering
    \begin{subfigure}[t]{0.45\textwidth}
        \includegraphics[width=\textwidth]{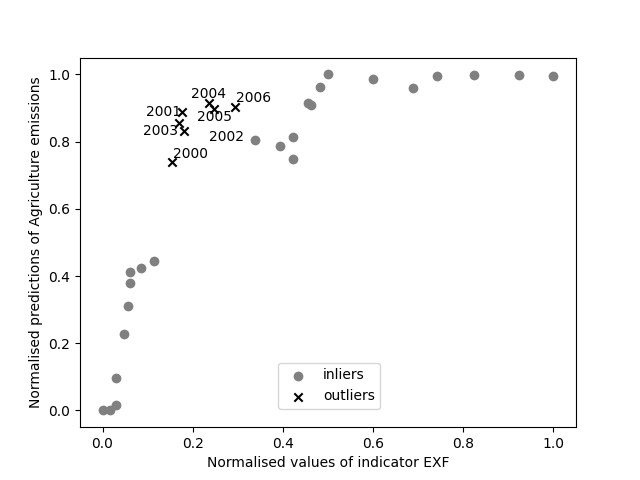}
        \caption{EXF: \EXF}
        \label{fig:indicator_fruit_and_veges}
    \end{subfigure}
    \hspace{1cm}
    \begin{subfigure}[t]{0.45\textwidth}
        \includegraphics[width=\textwidth]{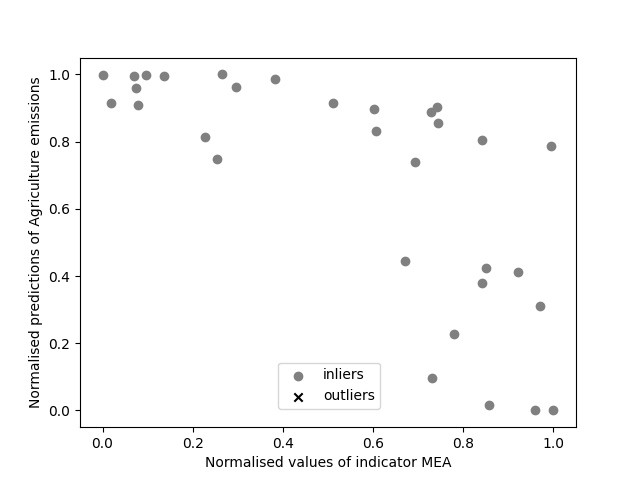}
        \caption{MEA: \MEA}
        \label{fig:indicator_NZMB}
    \end{subfigure}
    \caption{Inferred relationships between indicators and predicted emissions. Outliers are identified and labelled with their year.}
    \label{fig:sub_figures_indicators}
\end{figure}

The indicator EXF (\EXF) for Agriculture has a logarithmic trend (see \cref{fig:indicator_fruit_and_veges}). This is to be expected in horticulture since smaller farms are known to have higher average costs, which are associated to emissions. Thus, as production increases, we would expect emissions to increase at a high rate initially and then at a lower rate for higher levels of production. Outliers in this inferred relationship occurred between years 2001 and 2006. This coincides with a surge of Agriculture emissions. This surge is largely due to enteric fermentation in non-dairy cattle (category code 3.A.1 in the NZGGI~\cite{MfE_national_inventory_report_2020}). However, EXF is not related to non-dairy cattle, so it is natural that the indicator could have as low values as it does in spite of the surge.

All indicators had unsurprisingly positive inferred relationships with emissions except for the following example. On average, predicted Agriculture emissions decreased while MEA (\MEA) increased (see \cref{fig:indicator_NZMB}). Since every animal is a source of emissions, if animals are removed from the stock at a higher and higher rate (in order to support higher levels of meat production), this must lead to a reduction in emissions with all else being equal. Indeed other factors will come into play in order to support the continuous turnover of the stock. Recognising such other factors would be the role of other indicators like COW, which counts the total number of dairy cows.

Therefore, we stand by the negative relationship observed between MEA and emissions in the Agriculture sector. We also emphasise that we do not imply there is a direct causal relationship between animals' lives being ended and a decrease in emissions. Rather there is an explainable relationship that should be viewed not in isolation but in the context of other interactive factors in the model. In this way, the negative relationship between Agriculture emissions and meat production can be explained. No outliers were identified in this inferred relationship.

\subsection{Preliminary forecast of New Zealand Greenhouse Gas Inventory}

\begin{figure}
    \centering
    \includegraphics[width=0.8\linewidth]{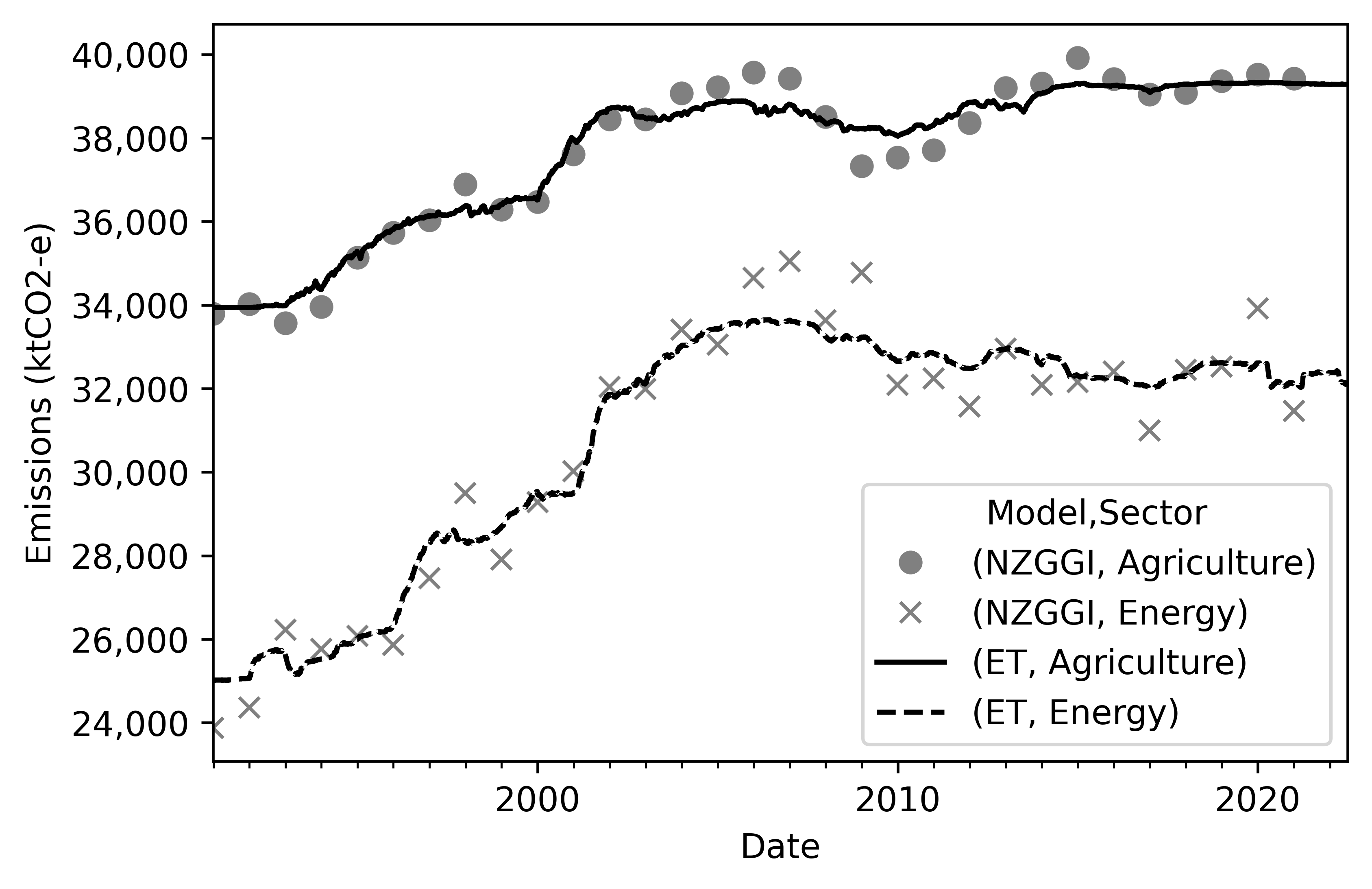}
    \caption{Emissions estimates for each sector modelled in our analysis according to NZGGI (the actual data) and extra trees--ET (the predicted data). This demonstrates the greater timeliness of our estimates both by being more frequent and having a lower latency (nowcasting).}%
    \label{fig:emissions_estimates}%
\end{figure}

We observe similar trends between our emissions estimates and the NZGGI (see \cref{fig:emissions_estimates}). However, our model underestimates the magnitude of peaks and troughs in the Agriculture sector and to a lesser degree in the Energy sector. This is a known weakness of machine learning in predicting extremes~\cite{MKSK22}.

We estimate Agriculture emissions in 2021 were $39,286$ ktCO\textsubscript{2}-e compared to $39,299$ ktCO\textsubscript{2}-e in 2020, which is a decrease of about $0.03\%$. We estimate Energy emissions in 2021 were $32,388$ ktCO\textsubscript{2}-e compared to $32,069$ ktCO\textsubscript{2}-e in 2020, which is an increase of about $0.99\%$.

\subsection{Energy emissions in 2020 COVID-19 lockdown}\label{sec:energy_and_covid}

We estimate total Energy emissions were $32,576$ ktCO\textsubscript{2}-e in 2019, higher than the 2020 figure above, but only by about $1.58\%$. This comes in spite of several COVID-19 lockdowns. Indicators in our model for Energy appeared to recover from their shocks too quickly for lockdowns to have lasting impact for the year of 2020.

\begin{figure}[H]
    \centering
    \includegraphics[width=0.8\linewidth]{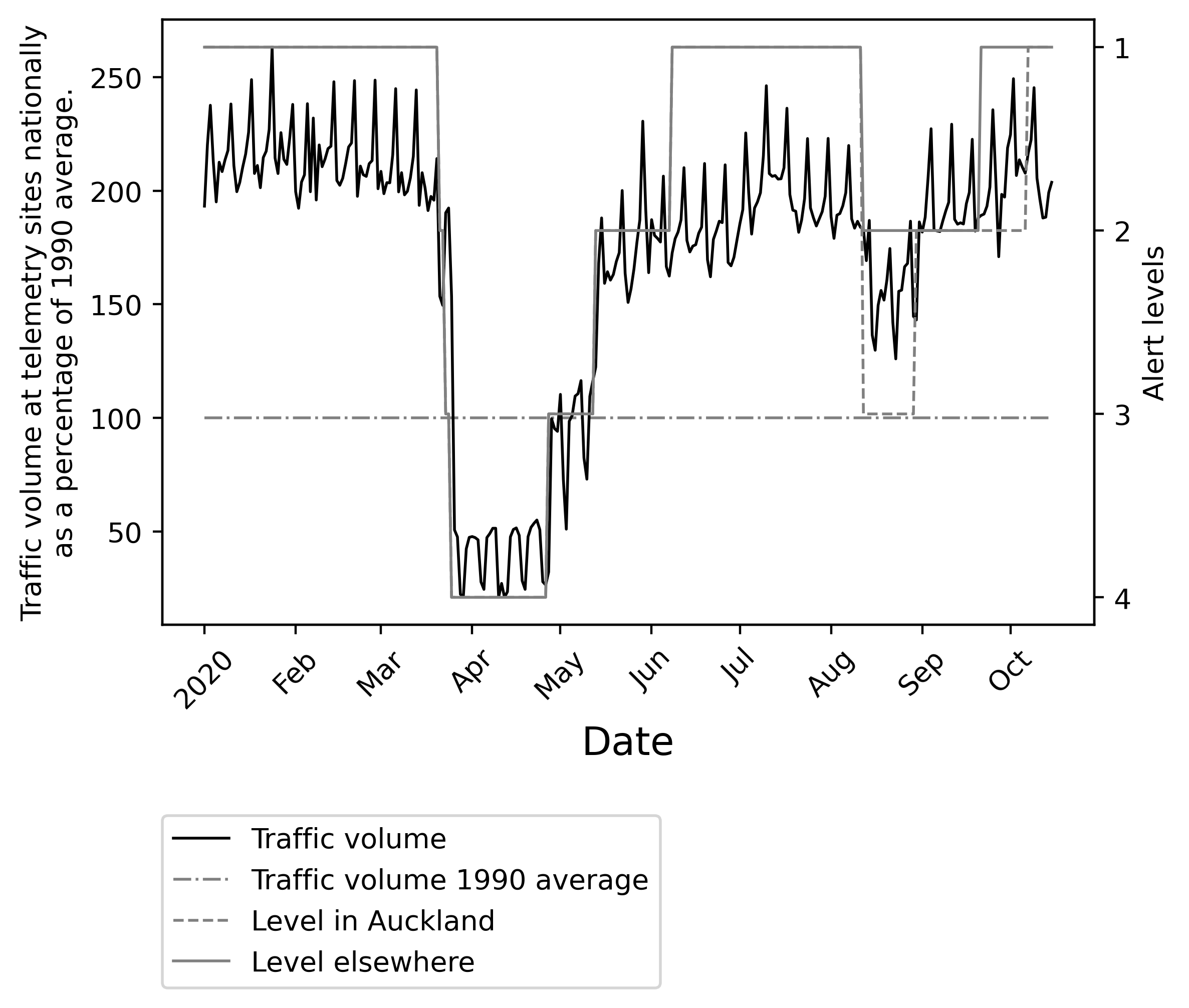}
    \caption{Overlay of COVID-19 alert levels in Auckland and elsewhere in New Zealand with daily national traffic volume at selected sites across the country. The traffic volume is expressed as a percentage of the 1990 average. It is involved in the following indicators: LEE (\LEE) and ETV (\ETV).}
    \label{fig:fleet_reduction_lockdown}
\end{figure}

For example, consider transport (which accounted for $41.88\%$ of Energy emissions in 2020~\cite{MfE_emissions_tracker}); changes in state highway traffic volumes at selected telemetry sites across the country--a component of indicators in our model for Energy--were closely related to changes in COVID-19 alert levels (see \cref{fig:fleet_reduction_lockdown}). Thus, traffic volumes recovered before the end of the year, as did emissions in the Energy sector.

\newpage
\subsection{Link between emissions and economic productivity}\label{sec:emissions_and_productivity}

Because we maintain our indicator data on a daily frequency, our model can make predictions on a daily basis of emissions occurring during 365 day windows, which aligns with the NZGGI on 31 December each year. An advantage this gives us is visualising more granular trends than if we only had annual predictions. We demonstrate the value of this through an example.

\begin{figure}[H]
    \centering
    \includegraphics[width=0.8\linewidth]{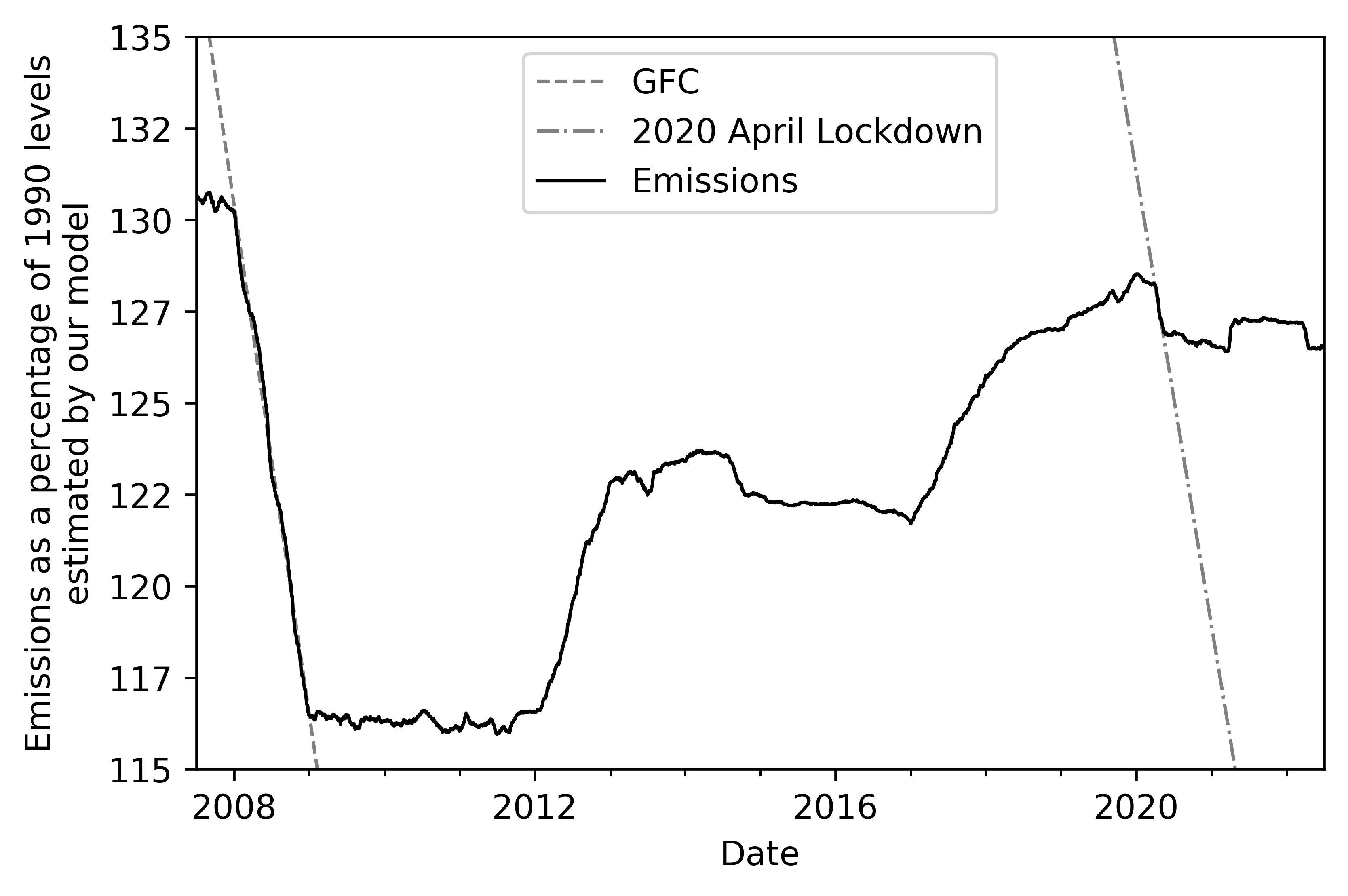}
    \caption{Net emissions as a percentage of 1990 levels with lines of best fit during the global financial crisis (GFC) and the COVID-19 lockdown from 25 March to 26 April 2020. This demonstrates the slope during the GFC (decrease of 1\% every 26.2 days) is similar to that of the 2020 April lockdown (decrease of 1\% every 29.5 days). In this sense, the GFC and 2020 April lockdown had comparable effects on net emissions.}
    \label{fig:gfc_comparison}
\end{figure}

An historical comparison to the Global Financial Crisis (GFC) suggests a continued link between emissions and economic productivity. Observe in \cref{fig:gfc_comparison} that the steep descent of emissions during the GFC (beginning in 2008) has a similar gradient to that of COVID-19 lockdown from 25 March to 26 April 2020. Given both the GFC and COVID-19 lockdowns are associated to significant economic impact, this raises the question of how exactly emissions is associated with economic activity.

The release of emissions budgets by the New Zealand Government in 2021 highlighted a turning point for the dissociation of economic activity and emissions. Budgets can be monitored closely with our dynamic estimates.

\section{Discussion}\label{sec:discussion}

\subsection{Strengths}

\subsubsection{Meeting the needs of nowcasting}\label{sec:needs_of_nowcasting}

The outputs of the model are daily predictions of cumulative emissions in 365 day periods per sector, which coincide with the national inventory at the end of each year. This comes with a time lag of about two months given our current indicator data. Therefore, our model addresses four of five needed directions of improvements to emissions statistics demanded in~\cite{nordic_nowcasting}: time lag, frequency, sector-level activity and accuracy (discussed in \cref{sec:accuracy} among our other model outcomes). The fifth direction is coverage of sources of emissions, which we discuss in \cref{sec:limiations_indicators}. Thus, we meet a growing demand for nowcasting (see references in \cref{sec:introduction}).

\subsubsection{Repeatability}

Our emissions estimates anchor to and complement the NZGGI. A key strength is repeatability for every new edition of the NZGGI. We hope this continuity of information on emissions will hold policies and actions to account.

\subsubsection{Comparison to methodologies using bespoke inventories}

Innovative techniques have been developed to investigate COVID-19 impacts on daily emissions \cite{forster2020current,liu2020carbon,liu2020near,oda2021errors,le2020temporary,zheng2020satellite} with impressive results (e.g. daily emissions with a regional component separating CO\textsubscript{2} and air pollutants~\cite{huang2021highly}). However, these studies are limited by depending on bespoke and labour-intensive daily inventories of emissions. To mitigate the labour, they compensate in various ways.

Firstly, the studies focus on short windows of time, whereas our work reports from 1990 up to two months prior to the present day. Greenhouse gas emissions systems change gradually, so our long-term perspective is valuable.

Secondly, scaling/repeating methods in these studies would be costly since it would require curating another bespoke inventory. In particular, scaling the methods to a national inventory like the NZGGI is hard to foresee; the 15 month delay before each edition of the NZGGI is released suggests the annual NZGGI is labour-intensive enough.

Lastly, the daily inventories used in these studies try to reproduce third-party estimates of countries' national emissions, rather than official estimates. This leads to a fragmented picture of emissions. Our methodology aligns with the existing national inventory.

\subsubsection{Local quarterly emissions reports}\label{sec:local quarterly emissions reports}

Statistics New Zealand (StatsNZ) has begun experimental reporting on quarterly emissions~\cite{SNZ_quarterly_emissions_example}. We comment on distinctions between our methodologies here.

StatsNZ reports are more dynamic in one way than ours, estimating emissions occurring within any given quarter, whereas our estimates are always for 365 day periods. However, our estimates are more dynamic than theirs in the sense that we provide daily predictions (of annual periods). Lastly, StatsNZ reports currently involve a different classification of `industries' compared to the `sectors' in the NZGGI. 

\subsubsection{Novelty of our approach}\label{sec:comparison_to_other_methods}

The review by~\cite{Deb21} of 776 articles on emissions predictions suggests tree-based models are not the most widespread. We did find a handful of studies using tree-based models for emissions predictions~\cite{CMW18,JPL22,MKSK22,LIN2021121502,,WEN2020137194,YANG2020124071}, but their applications vary from ours. Hence, our research is novel in terms of the application of tree-based models to estimating the NZGGI. Furthermore, the performance of our methods suggests that techniques that can robustly handle data that potentially violates the assumptions of regression are most suitable, such as random forest and extra trees.

The most similar studies to ours by methodology and application were~\cite{CMW18,JPL22,Ulk22}. We remark on points of difference that are strengths of our approach.

Re-training our model day after day is operationally necessary to incorporate the latest data and give the latest estimates. This raises a question of the stability of the structure of the model between training, which was assessed in \cref{sec:stability}. The comparison studies do not assess stability.

Our selection protocol from \cref{sec:final_selection} is decision-driven instead of quantitatively-driven as in~\cite[Figure 5]{CMW18}, which is more definitive but less flexible than ours. That is, \cite{CMW18} only accounts for average performance, whereas our protocol allows us to take into account the distribution of performance throughout simulations. This is important because model selection has its subtleties; regarding over-fitting, \cite{prevent_overfitting} provides evidence that finding the optimal model is not the same as finding the model with the least over-fitting. We discuss a blend of the protocols in \cref{sec:methodological_limitations}.

Recall our selection protocol is based on three model outcomes: accuracy, stability and over-fitting. Cross validation is used to gather information these outcomes. Compared to similar studies~\cite{JPL22,Ulk22}, our analysis and results:
\begin{enumerate}
    \item Employ a more thorough cross validation approach (\cref{sec:cross_validation}),
    \item Demonstrate higher accuracy (relative to contemporary methods referenced in \cref{sec:accuracy}), and
    \item Assess the extent of over-fitting (\cref{sec:overfitting}).
\end{enumerate}

\subsection{Limitations}

Limitations are considered in three types: 
\begin{itemize}
    \item Methodological limitations: ways we are limited due to our approach in theory,
    \item Operational limitations: limited availability of data, and
    \item Explanatory and predictive limitations: limitations on the how well the model explains the system at hand and on the predictions we can make with the model.
\end{itemize}

\subsubsection{Methodological limitations}\label{sec:methodological_limitations}

A key limitation of the methodology is that it treats high level aggregates directly by taking sectors' emissions as dependent variables. We could enhance the model by partitioning sectors whenever we find that a source can be modelled separately into `the source' and `not the source' parts. The extreme case of this approach would be to model every source and sink individually.

Also, the methodology estimates CO\textsubscript{2}-e variables directly. Modelling each greenhouse gas type is of interest from both a policy and practical perspective.

Our model estimates the NZGGI, which carries the imperfections of the IPCC guidelines it follows \cite{yona2020refining}. Thus, our model limited in every way the NZGGI is limited.

As described in \cref{sec:comparison_to_other_methods}, the protocol we use to select our preferred model is less definitive than in~\cite{CMW18}. In order to make our protocol more definitive but maintain a view of the variation of model performance between simulations, future research could investigate the following approach: define an overall performance score by computing differences between upper and lower quartiles for each model outcome (accuracy, stability and over-fitting), multiplying the difference by the mean of the model outcome when the model outcome is supposed to be minimised (accuracy and stability), and taking a weighted sum of these components. The optimal model for a given sector would be the model that minimises this score for that sector. If different sectors have different optimal models, the score could be broadened to aggregate over sectors in order to determine a general optimal model.

\subsubsection{Operational limitations}

Data availability is the main limitation to the operational quality of the model. In particular, limited availability leaves us with a latency of two months. For this reason we are always investigating new sources of dynamic and near-real-time data to improve the model.

\subsubsection{Explanatory and predictive limitations}

Whilst our model selection prioritises predictive power over explanatory power~\cite{Bre01}, we ensure that our models are explainable. To this end, we (1) selected indicators with known systematic relationships with emissions and (2) verified that the relationships between indicators and predictions were sensible. Future research will involve increasing the explanatory power of the model.

In order for prediction accuracy statistics in \cref{tab:test_mape} to apply to future estimates, we assume there is no change from one NZGGI edition to the next. We measure how bad an assumption this is by the extent of the changes between editions. For example, from the 2018 edition to the 2019 edition in the Agriculture sector the emissions for each year reported on changed by between $4.2\%$ and $5.2\%$. Whereas from the 2019 edition to the 2020 edition in the Energy sector, the reports changed by between $-0.5\%$ and $1.6\%$. This presents a moving target for our model. Estimating compound uncertainty due to these NZGGI amendments is non-trivial because amendments are non-uniform as noted above; reports for some years are amended to be higher while some years are amended to be lower. This dependence on the inventory at hand is a common disadvantage across any model estimating an inventory that updates over time. However, the advantage of using an updating inventory is the longevity of the applicability of our model. That is, since our model can be applied to any edition of the NZGGI, we continually revise our analysis according to new editions of the NZGGI each year.

\subsubsection{Indicator limitations}\label{sec:limiations_indicators}

The quality of any model depends on the quality of the data to which it is fit. Therefore, it is important to note limitations of our indicators. Their constructions are described in \cref{appendix:indicators}. They have a limited coverage of the many sources of emissions, and coverage is an important direction for improvement of emissions information~\cite{nordic_nowcasting}. Screening a larger number of indicators when data becomes available will be beneficial for the coverage of our model, and a formal indicator selection such as in \cite{WEN2020137194} will be necessary.

Many indicators do not account for emissions intensities. Multiplying indicator values by a coefficient from an IPCC look-up table is not useful because models tend to be mostly scale-independent. Thus, a time-dependent analysis of emissions intensity for the given activity is necessary.

We present a preliminary model, with a limited number of indicators covering only Energy and Agriculture. We recognise the importance of dynamically modelling the remaining sectors of the NZGGI. Fortunately, the majority of the work has been done, resulting in a generic sector model. Modelling the remaining sectors only requires trialling new data sources when they become available and assessing their outcomes.

\section{Conclusions}

This study develops a proof of concept that there is a framework by which machine learning can be used in combination with a small number of indicators to model sector-level emissions estimates at a higher frequency and lower latency to complement the New Zealand Greenhouse Gas Inventory (NZGGI) and any other relevant methodologies currently investigating greenhouse gas emissions in New Zealand. Estimates are reported every day for the cumulative emissions in the 365 day period ending that day, which gives a more continuous stream of emissions statistics with a time lag of approximately two months.

The high frequency allows us to compare shocks that take effect in time periods shorter than one year. For example, our results show the COVID-19 lockdown from 25 March to 26 April 2020 in New Zealand coincided with a rate of decrease in emissions similar to that of the global financial crisis of 2008. 

The low latency allows us to forecast, up to 2021, the latest edition of the NZGGI (reporting only up to 2020) to estimate total emissions in the Agriculture and Energy sectors. We estimate Agriculture emissions in 2021 were $39,286$ ktCO\textsubscript{2}-e and Energy emissions in 2021 were $32,388$ ktCO\textsubscript{2}-e, which represents a decrease of $0.03\%$ and an increase of $0.99\%$, respectively, since 2020.

In the face of rapid policy and technological change, these insights based on our model can improve current understanding of emissions in a landscape changing day to day.

We evaluate uncertainties in our methodology and performance of a number of models in terms of prediction accuracy, stability, over-fitting, and by comparison to typical inter-annual variation. This assessment is more rigorous than contemporary studies in many ways. The selected model carries an error margin of $1.314\%$ for Agriculture and $3.194\%$ for Energy (median during simulation study). The model is explainable in the sense that indicators used have systematic relationships with the emissions in their sectors, and the model fits sensible quantitative relationships between each indicator and the predictions it makes. That said, the model is not designed to tell us which indicators of emissions are stronger drivers than others.

This study has highlighted the demand for the release of relevant data sets in a more timely fashion. More granular data would lead to significant improvements to the model. We also emphasise the need for the continuation of a bottom-up emissions model as presented in national greenhouse gas inventories; this would provide a test-bed for a more dynamic approach and to directly measure the role of technological advances across the broad spectrum of emissions.

We note that although methodologies provided in this paper were able to estimate sub annual inventory style estimates within New Zealand, transferring this methodology to other countries would prove challenging, due to the nature of data used in this analysis.

Our methodology and emissions estimates will continue to be updated and developed to assess the ongoing variability in New Zealand's emissions. Further research will be undertaken for various purposes as discussed in \cref{sec:discussion}, not least of which in order to estimate emissions by gas type, providing direct comparison against progress towards split gas targets, and regional emissions estimates that support more direct action at a local level. We also plan to expand our estimates to other sectors of the NZGGI, although note that our current methodology dynamically reports on sectors making up 90.0\% of New Zealand's emissions.

\section*{Declaration of competing interest}

The authors declare that they have no known competing financial interests or personal relationships that could have appeared to influence the work reported in this paper.

\section*{Data availability}

The data used in this research is either publicly available via the cited source or the data was made available to the researchers through a data commercialisation agreement with one of the data partners of DOT Loves Data. The data provided by the data commercialisation agreement is therefore unavailable to access.

\section*{Acknowledgements}

We would like to thank the wider DOT Loves Data team for their ongoing support and advice during the development of the Dynamic Carbon Tracker, especially Moritz Lenz for important guidance in the literature and methodological support. We would also like to thank Louisa Howse for their suggestions in the early stages of method development, along with those at Transpower and emsTradepoint for their continued interest. The project has been jointly funded by Transpower, Callaghan Innovation and DOT Loves Data.

\appendix
    
\section{Indicators}\label{appendix:indicators}

\cref{tab:indicators} collates details of indicators used as independent variables in sector models. Descriptions of these indicators are given in the following subsections.

\subsection{\LEE}\label{sec:liquid_and_electric_energy}

This indicator comprises of an aggregate of traffic volumes and the estimated carbon emissions due to national electricity generation~\cite{em621}. Each component is described below. The increasing rate of electric vehicles draws a systematic connection between the components, which partly motivates their aggregation.

The components are normalised and added together with equal weight to give `\LEE'.\footnote{This aggregation was undertaken initially to avoid collinear indicators when we were considering only models that demanded this. Since the introduction of the tree-based models, it is now feasible to remove this aggregate, which we will undertake in future research.} Equal weighting is chosen to give the model equal view of both variables since we are interested in trends rather than magnitude of sources of emissions.

\textbf{Traffic volumes variable.} Waka Kotahi New Zealand Transport Agency provides an application programming interface for daily traffic volumes at sites around the country~\cite{NZTA21b} dating back to 2018. We restrict our attention to telemetry sites because of their accuracy and frequency. We extend the data back to 1990 by matching it with annual historic records in pdfs and xlsxs~\cite{NZTA21c}. This gives a unified record of traffic volumes with many site descriptions. A difficulty is that the description of a given site can change over time. To connect the traffic volumes per site in spite of changing descriptions, we built an adjacency table between the site descriptions and a manual selection of keywords (e.g. `Ngauranga', `Drury' and `Taupiri') where a keyword is adjacent to a site description if the key word appears in the site description. Sites being established or disestablished over time could lead to over- or under-estimates, respectively, of the increase in traffic volumes over time. For this reason we restricted to only those sites that have run continuously from 1990 to the present day. The traffic volume variable is the time series given by summing the series for each of these sites.

\textbf{Electricity generation variable.} The estimated carbon emissions due to national electricity generation is defined in the em6 application programming interface integration guide~\cite{em621}.

\subsection{\REM}\label{sec:newly_registered_emissions}

This indicator captures the emissions potential of New Zealand's vehicles based on motive power, year of manufacture and gross vehicle mass. Previous studies have highlighted the importance of including a view of the country's vehicle fleet~\cite{GSMO20}.

Vehicles in the Motor Vehicle Register~\cite{NZTA21a} are assigned an emissions potential (kgCO\textsubscript{2}-e/km) based on locally-approved IPCC-developed emissions factors~\cite{MfE_detailed_guide}. The emissions factors consider motive power (`petrol', `diesel', `petrol hybrid', `electric', `lpg', `plugin petrol hybrid', `petrol electric hybrid') and vehicle year, assuming a gross vehicle mass of 2000-3000kg. We assign to each vehicle the relevant emissions factor after adjusting for its actual gross vehicle mass. The adjustment applies the principle that fuel economy is proportional to vehicle mass, which is comparable to the proportions-based approach to understanding environmental impact of vehicles in terms of their mass~\cite{del2017effect}. Other studies use a congestion index~\cite{liu2020near} and may match traffic flow data with a vehicle database to differentiate gasoline from diesel powered vehicles~\cite{huang2021highly}. Our view has an advantage over these methodologies by differentiating vehicles by more motive power categories as well as by vehicle age and gross vehicle mass.

\subsection{\ETV}\label{sec:ETV}

This indicator is the product of the emissions potential of the fleet and the traffic volumes variable from \cref{sec:liquid_and_electric_energy}. Estimating the emissions potential of the fleet involved estimating the half-life of a vehicle on New Zealand roads and using this in an exponentially weighted mean over newly registered vehicles each carrying their own emissions potential.

Recall our indicator of emissions intensity of vehicles registered onto New Zealand roads from \cref{sec:newly_registered_emissions} based on the New Zealand Motor Vehicle Register (MVR). A limitation of this indicator is that we do not know how long these vehicles spend on the road. There does exist a dashboard tool that provides a `vehicle status' feature for the MVR at an aggregated level, which tells whether a vehicle is currently active or inactive. However, the API that provides the MVR to us vehicle-by-vehicle does not include `vehicle status'. Therefore, this feature was not readily available to us in order to calculate how long vehicles spend on the road. Instead we opted to use more readily available data for this purpose.

Our estimate of the half-life of a vehicle is as follows. We record how long the vehicles were not on New Zealand roads by taking the mean difference in features from the MVR: `vehicle year' and `first NZ registration year'. We denote this mean difference by $YBF$ for `years before registration'. Next a local scrappage report~\cite{scrappage} can indicate typical ages of vehicles when they are de-registered, which reported an average of 16.7 years and 18.6 years in Wellington and Christchurch, respectively. Thus, we take $\text{mean}\{16.7,18.6\} - YBF$ to be the half-life of vehicles on New Zealand roads. Then, we use the above half-life to take the exponentially weighted moving mean of the `\REM' indicator. This constitutes an exponential decay model in terms of the average lifespan of vehicles. There is a precedent for this in the literature, for example~\cite[Equation (1)]{Oguchi2015}. This gives a variable indicative of the fleet's emissions potential over time.

The product of the fleet emissions potential and the (state highway) traffic volumes variable forms this indicator.

The half-life used above is a rough estimate sufficient only for this proof of concept. In future research we plan to investigate refining our estimate of half-life. This could involve accessing the data in the MVR behind the dashboard mentioned above that provides `vehicle status'. As well as providing a national view of vehicle activity, this could be updated whenever the MVR is updated, which would be more dynamic that above where the half-life estimate is deduced from a historic scrappage report~\cite{scrappage}.

\subsection{\COA}

We extract quarterly values of total coal production from `Table 1 - Quarterly Tonnes' of the `Quarterly Coal Supply, Transformation, \& Consumption (Tonnes)' publicly available spreadsheet `Data tables for coal'~\cite{MBIE21}.

Coal is used on a significant scale for energy in the metal, mineral, chemical and manufacturing industries, and construction~\cite{MfE_national_inventory_report_2019}. Moreover, an increase in coal-fired electricity generation drove an increase in public electricity and heat production emissions between 2018 and 2019~\cite{MfE_national_inventory_report_2019}.

\subsection{\COW}

We extract annual values for `Total Dairy Cattle (including Bobby Calves)' from \cite{AGR001AA}. In 2020, enteric fermentation by dairy cattle made up $35.6\%$ of emissions in the Agriculture sector and has increased by $65.5\%$ since 1990~\cite{MfE_emissions_tracker}. This increase is primarily driven by increases in dairy cattle numbers \cite[Page 11]{MfE_national_inventory_report_2019}, which is the purpose of this indicator.

\subsection{\MEA}\label{sec:MEA}

We collate weekly weight statistics of beef and sheep meat production for export reported by the New Zealand Meat Board from certain export-inspected processing facilities dating back to the 2017-18 season. We sum the weight of all types of meat produced across the country. The series is retrofitted back to 1990 with monthly values of `Total Above Livestock Excluding Game' from \cite{LSS025AA}.

\subsection{\FIM}

We extract monthly values for the quantity of imported `Fertilisers' as in \cite{IMP033AA}. Use of synthetic nitrogen fertilisers is a driver of emissions trends in Agriculture \cite[Page 71]{MfE_national_inventory_report_2019}. The use of fertiliser consumption as an indicator has been effective in~\cite{IAS22}.

\subsection{Food exports}

Food export data is used as a general measure of food production. The indicators `\EXM', `\EXD' and `\EXF' are given by extracting monthly values for `Live Animals, Meat and Edible Meat Offal', `Fish, Crustaceans, Molluscs, Dairy Produce and Other Animal Products', and `Vegetables, Fruit and Prepared Foodstuffs, Beverages and Tobacco' respectively, from \cite{EXP012AA}. We acknowledge that that this is a price-based indicator, and that price variations as well as production amounts are therefore included. Further development would be to inflation-adjust this indicator to isolate the production component, or replace it entirely with a mass-based food export dataset.

\printbibliography

@misc{AGR001AA,
  author       = {{[dataset] Statistics New Zealand}},
  title        = {{AGR001AA} ({I}nfoshare)},
  howpublished = {\url{http://infoshare.stats.govt.nz/}},
  year         = 2022
}

@article{BB12,
  author  = {Bergstra, J and Bengio, Y},
  title   = {Random Search for Hyper-Parameter Optimization},
  journal = {Journal of Machine Learning Research},
  year    = {2012},
  volume  = {13},
  number  = {10},
  pages   = {281--305},
  url     = {http://jmlr.org/papers/v13/bergstra12a.html}
}

@article{Bre01,
  doi       = {10.1214/ss/1009213726},
  url       = {https://doi.org/10.1214/ss/1009213726},
  year      = {2001},
  month     = aug,
  publisher = {Institute of Mathematical Statistics},
  volume    = {16},
  number    = {3},
  author    = {Leo Breiman},
  title     = {Statistical Modeling: The Two Cultures (with comments and a rejoinder by the author)},
  journal   = {Statistical Science}
}

@article{CMW18,
  doi       = {10.1016/j.buildenv.2018.09.054},
  url       = {https://doi.org/10.1016/j.buildenv.2018.09.054},
  year      = {2018},
  month     = dec,
  publisher = {Elsevier {BV}},
  volume    = {146},
  pages     = {238--246},
  author    = {Shisheng Chen and Kuniaki Mihara and Jianxiu Wen},
  title     = {Time series prediction of {CO}2,  {TVOC} and {HCHO} based on machine learning at different sampling points},
  journal   = {Building and Environment}
}

@article{Deb21,
  doi       = {10.1016/j.uclim.2021.100849},
  url       = {https://doi.org/10.1016/j.uclim.2021.100849},
  year      = {2021},
  month     = may,
  publisher = {Elsevier {BV}},
  volume    = {37},
  pages     = {100849},
  author    = {Daniela Debone and Vinicius Pazini Leite and Simone Georges El Khouri Miraglia},
  title     = {Modelling approach for carbon emissions,  energy consumption and economic growth: A systematic review},
  journal   = {Urban Climate}
}

@article{del2017effect,
  doi       = {10.1016/j.jclepro.2017.04.013},
  url       = {https://doi.org/10.1016/j.jclepro.2017.04.013},
  year      = {2017},
  month     = jun,
  publisher = {Elsevier {BV}},
  volume    = {154},
  pages     = {566--577},
  author    = {Francesco Del Pero and Massimo Delogu and Marco Pierini},
  title     = {The effect of lightweighting in automotive {LCA} perspective: Estimation of mass-induced fuel consumption reduction for gasoline turbocharged vehicles},
  journal   = {Journal of Cleaner Production}
}

@misc{em621,
  author       = {{[dataset] Energy Market Services}},
  title        = {Current Carbon Intensity {API} (em6 {API})},
  howpublished = {\url{https://www.ems.co.nz/em6-api-integration-guide/}},
  year         = 2022,
  note         = {Version 1.9}
}

@misc{EXP012AA,
  author       = {{[dataset] Statistics New Zealand}},
  title        = {{EXP012AA} ({I}nfoshare)},
  howpublished = {\url{http://infoshare.stats.govt.nz/}},
  year         = 2022
}

@article{farrar1967multicollinearity,
  doi       = {10.2307/1937887},
  url       = {https://doi.org/10.2307/1937887},
  year      = {1967},
  month     = feb,
  publisher = {{JSTOR}},
  volume    = {49},
  number    = {1},
  pages     = {92},
  author    = {Donald E. Farrar and Robert R. Glauber},
  title     = {Multicollinearity in Regression Analysis: The Problem Revisited},
  journal   = {The Review of Economics and Statistics}
}

@article{FLV09,
  doi       = {10.1002/cem.1225},
  url       = {https://doi.org/10.1002/cem.1225},
  year      = {2009},
  month     = apr,
  publisher = {Wiley},
  volume    = {23},
  number    = {4},
  pages     = {160--171},
  author    = {Peter Filzmoser and Bettina Liebmann and Kurt Varmuza},
  title     = {Repeated double cross validation},
  journal   = {Journal of Chemometrics}
}

@article{forster2020current,
  doi       = {10.1038/s41558-020-0883-0},
  url       = {https://doi.org/10.1038/s41558-020-0883-0},
  year      = {2020},
  month     = aug,
  publisher = {Springer Science and Business Media {LLC}},
  volume    = {10},
  number    = {10},
  pages     = {913--919},
  author    = {Piers M. Forster and Harriet I. Forster and Mat J. Evans and Matthew J. Gidden and Chris D. Jones and Christoph A. Keller and Robin D. Lamboll and Corinne Le Qu{\'{e}}r{\'{e}} and Joeri Rogelj and Deborah Rosen and Carl-Friedrich Schleussner and Thomas B. Richardson and Christopher J. Smith and Steven T. Turnock},
  title     = {Current and future global climate impacts resulting from {COVID}-19},
  journal   = {Nature Climate Change}
}

@article{GEW06,
  doi       = {10.1007/s10994-006-6226-1},
  url       = {https://doi.org/10.1007/s10994-006-6226-1},
  year      = {2006},
  month     = mar,
  publisher = {Springer Science and Business Media {LLC}},
  volume    = {63},
  number    = {1},
  pages     = {3--42},
  author    = {Pierre Geurts and Damien Ernst and Louis Wehenkel},
  title     = {Extremely randomized trees},
  journal   = {Machine Learning}
}

@article{Ghorbani2019,
  doi       = {10.22190/fumi1903583g},
  url       = {https://doi.org/10.22190/fumi1903583g},
  year      = {2019},
  month     = oct,
  publisher = {University of Nis},
  pages     = {583},
  author    = {{H}amid {G}horbani},
  title     = {Mahalanobis distance and its application for detecting multivariate outliers},
  journal   = {Facta Universitatis,  Series: Mathematics and Informatics}
}

@article{GSMO20,
  doi       = {10.3390/su12031012},
  url       = {https://doi.org/10.3390/su12031012},
  year      = {2020},
  month     = jan,
  publisher = {{MDPI} {AG}},
  volume    = {12},
  number    = {3},
  pages     = {1012},
  author    = {Mariano Gonz{\'{a}}lez-S{\'{a}}nchez and Juan Luis Mart{\'{i}}n-Ortega},
  title     = {Greenhouse Gas Emissions Growth in Europe: A Comparative Analysis of Determinants},
  journal   = {Sustainability}
}

@article{huang2021highly,
  doi       = {10.1021/acs.estlett.1c00600},
  url       = {https://doi.org/10.1021/acs.estlett.1c00600},
  year      = {2021},
  month     = sep,
  publisher = {American Chemical Society ({ACS})},
  volume    = {8},
  number    = {10},
  pages     = {853--860},
  author    = {Cheng Huang and Jingyu An and Hongli Wang and Qizhen Liu and Junjie Tian and Qian Wang and Qingyao Hu and Rusha Yan and Yin Shen and Yusen Duan and Qingyan Fu and Jiandong Shen and Hui Ye and Ming Wang and Chong Wei and Yafang Cheng and Hang Su},
  title     = {Highly Resolved Dynamic Emissions of Air Pollutants and Greenhouse Gas {CO}2 during {COVID}-19 Pandemic in East China},
  journal   = {Environmental Science \& Technology Letters}
}

@incollection{Huang2022,
  doi       = {10.1007/978-981-19-1625-0_7},
  url       = {https://doi.org/10.1007/978-981-19-1625-0_7},
  year      = {2022},
  publisher = {Springer Nature Singapore},
  pages     = {237--254},
  author    = {Yue Huang},
  title     = {Linear Calibration Methods},
  booktitle = {Chemometric Methods in Analytical Spectroscopy Technology}
}

@article{IAS22,
  doi       = {10.21601/ejosdr/12176},
  url       = {https://doi.org/10.21601/ejosdr/12176},
  year      = {2022},
  month     = jun,
  publisher = {Modestum Ltd},
  volume    = {6},
  number    = {4},
  pages     = {em0194},
  author    = {Chukwuemeka Amaefule and Igwe J. Ibeabuchi and Akeem Shoaga},
  title     = {Determinants of Greenhouse Gas Emissions},
  journal   = {European Journal of Sustainable Development Research}
}

@misc{IMP033AA,
  author       = {{[dataset] Statistics New Zealand}},
  title        = {{IMP033AA} ({I}nfoshare)},
  howpublished = {\url{http://infoshare.stats.govt.nz/}},
  year         = 2022
}

@misc{Jamsranjav17,
  year         = 2019,
  author       = {E. {Calvo Buendia} and K. Tanabe and A. Kranjc and J. Baasansuren and M. Fukuda and S. Ngarize and A. Osako and Y. Pyrozhenko and P. Shermanau and S. Federici},
  title        = {2019 Refinement to the 2006 IPCC Guidelines for National Greenhouse Gas Inventories},
  howpublished = {\url{https://www.ipcc-nggip.iges.or.jp/public/2019rf/index.html}},
  note         = {Published: IPCC, Switzerland}
}

@inproceedings{JPL22,
  doi       = {10.1109/icoei53556.2022.9777119},
  url       = {https://doi.org/10.1109/icoei53556.2022.9777119},
  year      = {2022},
  month     = apr,
  publisher = {{IEEE}},
  author    = {Vaishnavi Jayaraman and Saravanan Parthasarathy and Arun Raj Lakshminarayanan},
  title     = {Forecasting the Emission of Greenhouse Gases from the Waste using {SARIMA} Model},
  booktitle = {2022 6th International Conference on Trends in Electronics and Informatics ({ICOEI})}
}

@article{le2020temporary,
  doi       = {10.1038/s41558-020-0797-x},
  url       = {https://doi.org/10.1038/s41558-020-0797-x},
  year      = {2020},
  month     = may,
  publisher = {Springer Science and Business Media {LLC}},
  volume    = {10},
  number    = {7},
  pages     = {647--653},
  author    = {Corinne Le Qu{\'{e}}r{\'{e}} and Robert B. Jackson and Matthew W. Jones and Adam J. P. Smith and Sam Abernethy and Robbie M. Andrew and Anthony J. De-Gol and David R. Willis and Yuli Shan and Josep G. Canadell and Pierre Friedlingstein and Felix Creutzig and Glen P. Peters},
  title     = {Temporary reduction in daily global {CO}2 emissions during the {COVID}-19 forced confinement},
  journal   = {Nature Climate Change}
}

@article{LIN2021121502,
  doi       = {10.1016/j.energy.2021.121502},
  url       = {https://doi.org/10.1016/j.energy.2021.121502},
  year      = {2021},
  month     = dec,
  publisher = {Elsevier {BV}},
  volume    = {236},
  pages     = {121502},
  author    = {Jinyao Lin and Siyan Lu and Xiaoyu He and Fang Wang},
  title     = {Analyzing the impact of three-dimensional building structure on {CO}2 emissions based on random forest regression},
  journal   = {Energy}
}

@article{liu2020carbon,
  doi       = {10.1038/s41597-020-00708-7},
  url       = {https://doi.org/10.1038/s41597-020-00708-7},
  year      = {2020},
  month     = nov,
  publisher = {Springer Science and Business Media {LLC}},
  volume    = {7},
  number    = {1},
  author    = {Zhu Liu and Philippe Ciais and Zhu Deng and Steven J. Davis and Bo Zheng and Yilong Wang and Duo Cui and Biqing Zhu and Xinyu Dou and Piyu Ke and Taochun Sun and Rui Guo and Haiwang Zhong and Olivier Boucher and Fran{\c{c}}ois-Marie Br{\'{e}}on and Chenxi Lu and Runtao Guo and Jinjun Xue and Eulalie Boucher and Katsumasa Tanaka and Fr{\'{e}}d{\'{e}}ric Chevallier},
  title     = {Carbon Monitor,  a near-real-time daily dataset of global {CO}2 emission from fossil fuel and cement production},
  journal   = {Scientific Data}
}

@article{liu2020near,
  doi       = {10.1038/s41467-020-18922-7},
  url       = {https://doi.org/10.1038/s41467-020-18922-7},
  year      = {2020},
  month     = oct,
  publisher = {Springer Science and Business Media {LLC}},
  volume    = {11},
  number    = {1},
  author    = {Zhu Liu and Philippe Ciais and Zhu Deng and Ruixue Lei and Steven J. Davis and Sha Feng and Bo Zheng and Duo Cui and Xinyu Dou and Biqing Zhu and Rui Guo and Piyu Ke and Taochun Sun and Chenxi Lu and Pan He and Yuan Wang and Xu Yue and Yilong Wang and Yadong Lei and Hao Zhou and Zhaonan Cai and Yuhui Wu and Runtao Guo and Tingxuan Han and Jinjun Xue and Olivier Boucher and Eulalie Boucher and Fr{\'{e}}d{\'{e}}ric Chevallier and Katsumasa Tanaka and Yiming Wei and Haiwang Zhong and Chongqing Kang and Ning Zhang and Bin Chen and Fengming Xi and Miaomiao Liu and Fran{\c{c}}ois-Marie Br{\'{e}}on and Yonglong Lu and Qiang Zhang and Dabo Guan and Peng Gong and Daniel M. Kammen and Kebin He and Hans Joachim Schellnhuber},
  title     = {Near-real-time monitoring of global {CO}2 emissions reveals the effects of the {COVID}-19 pandemic},
  journal   = {Nature Communications}
}

@misc{LSS025AA,
  author       = {{[dataset] Statistics New Zealand}},
  title        = {{LSS025AA} ({I}nfoshare)},
  howpublished = {\url{http://infoshare.stats.govt.nz/}},
  year         = 2022
}

@misc{MBIE21,
  author       = {{[dataset] Ministry of Business, Innovation and Employment}},
  title        = {Data tables for coal ({E}nergy statistics)},
  howpublished = {\url{https://www.mbie.govt.nz/building-and-energy/energy-and-natural-resources/energy-statistics-and-modelling/energy-statistics/coal-statistics/}},
  year         = 2022
}

@misc{MfE_detailed_guide,
  author       = {{Ministry for the Environment}},
  title        = {Measuring Emissions: A Guide for Organisations: 2020 Detailed Guide. Wellington: Ministry for the Environment.},
  howpublished = {\url{https://environment.govt.nz/assets/Publications/Files/Measuring-Emissions-Detailed-Guide-2020.pdf}},
  note         = {Accessed 12 Oct. 2022.},
  year         = 2022
}

@misc{MfE_emissions_tracker,
  author       = {{Ministry for the Environment}},
  title        = {New Zealand's Interactive Emissions Tracker},
  howpublished = {\url{https://emissionstracker.mfe.govt.nz/}},
  note         = {Accessed 8 Nov. 2021.},
  year         = 2022
}

@misc{MfE_national_inventory_report_2019,
  year         = 2020,
  author       = {{Ministry for the Environment}},
  title        = {New Zealand's Greenhouse Gas Inventory 1990–2019, National inventory report. Wellington: Ministry for the Environment},
  howpublished = {\url{https://environment.govt.nz/assets/Publications/Greenhouse-Gas-Inventory-1990-2019/New-Zealands-Greenhouse-Gas-Inventory-1990-2019-Volume-1-Chapters-1-15.pdf}},
  note         = {Accessed 13 Oct. 2022.}
}

@misc{MfE_national_inventory_report_2020,
  year         = 2021,
  author       = {{Ministry for the Environment}},
  title        = {New Zealand's Greenhouse Gas Inventory 1990–2020, National inventory report. Wellington: Ministry for the Environment},
  howpublished = {\url{https://environment.govt.nz/assets/publications/GhG-Inventory/New-Zealand-Greenhouse-Gas-Inventory-1990-2020-Chapters-1-15.pdf}},
  note         = {Accessed 13 Oct. 2022.}
}

@misc{MfE_spreadsheet,
  author       = {{[dataset] Ministry for the Environment}},
  title        = {Time series emissions data by category ({N}ational inventory report)},
  howpublished = {\url{https://environment.govt.nz/publications/new-zealands-greenhouse-gas-inventory-1990-2020/}},
  year         = 2022,
  note         = {ME 1635}
}

@article{MKSK22,
  doi       = {10.1002/saj2.20429},
  url       = {https://doi.org/10.1002/saj2.20429},
  year      = {2022},
  month     = jul,
  publisher = {Wiley},
  volume    = {86},
  number    = {5},
  pages     = {1227--1240},
  author    = {Ali Mehmandoost Kotlar and Jasdeep Singh and Sandeep Kumar},
  title     = {Prediction of greenhouse gas emissions from agricultural fields with and without cover crops},
  journal   = {Soil Science Society of America Journal}
}

@article{Mut22,
  doi       = {10.1007/s11356-022-20615-1},
  url       = {https://doi.org/10.1007/s11356-022-20615-1},
  year      = {2022},
  month     = may,
  publisher = {Springer Science and Business Media {LLC}},
  volume    = {29},
  number    = {45},
  pages     = {68332--68356},
  author    = {Mihai Mutascu},
  title     = {{CO}2 emissions in the {USA}: new insights based on {ANN} approach},
  journal   = {Environmental Science and Pollution Research}
}

@book{nordic_nowcasting,
  doi       = {10.6027/temanord2022-537},
  url       = {https://doi.org/10.6027/temanord2022-537},
  year      = {2022},
  month     = aug,
  publisher = {Nordisk Ministerr{\aa}d},
  author    = {Sofie Pandis Iveroth and David McKinnon and Jouni Tuomisto and Martin Wetterstedt and Agneta Persson},
  title     = {Nowcasting {CO}2 emissions}
}

@misc{NZMB21,
  author       = {{[dataset] New Zealand Meat Board}},
  title        = {Provisional Export Livestock Processing Data},
  howpublished = {\url{https://www.nzmeatboard.org/the-industry/production-data/}},
  year         = 2022
}

@misc{NZTA21a,
  author       = {{[dataset] New Zealand Transport Agency}},
  title        = {Motor Vehicle Register (Waka Kotahi {O}pen {D}ata)},
  howpublished = {\url{https://nzta.govt.nz/resources/new-zealand-motor-vehicle-register-statistics/new-zealand-vehicle-fleet-open-data-sets/\#data}},
  year         = 2022
}

@misc{NZTA21b,
  author       = {{[dataset] New Zealand Transport Agency}},
  title        = {{TMS} daily traffic counts API (Waka Kotahi {O}pen {D}ata)},
  howpublished = {\url{https://opendata-nzta.opendata.arcgis.com/datasets/tms-daily-traffic-counts-api/api}},
  year         = 2022
}

@misc{NZTA21c,
  author       = {{[dataset] New Zealand Transport Agency}},
  title        = {Traffic data booklets and state highway traffic volumes (Waka Kotahi {O}pen {D}ata)},
  howpublished = {\url{https://www.nzta.govt.nz/resources/state-highway-traffic-volumes/}},
  year         = 2019
}

@article{oda2021errors,
  doi       = {10.1088/1748-9326/ac109d},
  url       = {https://doi.org/10.1088/1748-9326/ac109d},
  year      = {2021},
  month     = aug,
  publisher = {{IOP} Publishing},
  volume    = {16},
  number    = {8},
  pages     = {084058},
  author    = {Tomohiro Oda and Chihiro Haga and Kotaro Hosomi and Takanori Matsui and Rostyslav Bun},
  title     = {Errors and uncertainties associated with the use of unconventional activity data for estimating {CO}2 emissions: the case for traffic emissions in Japan},
  journal   = {Environmental Research Letters}
}

@article{Oguchi2015,
  doi       = {10.1021/es505245q},
  url       = {https://doi.org/10.1021/es505245q},
  year      = {2015},
  month     = jan,
  publisher = {American Chemical Society ({ACS})},
  volume    = {49},
  number    = {3},
  pages     = {1738--1743},
  author    = {Masahiro Oguchi and Masaaki Fuse},
  title     = {Regional and Longitudinal Estimation of Product Lifespan Distribution: A Case Study for Automobiles and a Simplified Estimation Method},
  journal   = {Environmental Science \& Technology}
}

@inproceedings{prevent_overfitting,
  title     = {Preventing ``overfitting'' of cross-validation data},
  author    = {Ng, Andrew Y},
  booktitle = {ICML},
  volume    = {97},
  pages     = {245--253},
  year      = {1997},
  doi       = {10.5555/645526.657119}
}

@book{Python3,
  author    = {Van Rossum, Guido and Drake, Fred L.},
  title     = {Python 3 Reference Manual},
  year      = {2009},
  isbn      = {1441412697},
  publisher = {CreateSpace},
  address   = {Scotts Valley, CA}
}

@article{RS21,
  doi       = {10.1088/1748-9326/ac02ec},
  url       = {https://doi.org/10.1088/1748-9326/ac02ec},
  year      = {2021},
  month     = jun,
  publisher = {{IOP} Publishing},
  volume    = {16},
  number    = {6},
  pages     = {068002},
  author    = {Joeri Rogelj and Carl-Friedrich Schleussner},
  title     = {Reply to Comment on `Unintentional unfairness when applying new greenhouse gas emissions metrics at country level'},
  journal   = {Environmental Research Letters}
}

@article{SciPy,
  author  = {Virtanen, Pauli and Gommers, Ralf and Oliphant, Travis E. and Haberland, Matt and Reddy, Tyler and Cournapeau, David and Burovski, Evgeni and Peterson, Pearu and Weckesser, Warren and Bright, Jonathan and {van der Walt}, St{\'e}fan J. and Brett, Matthew and Wilson, Joshua and Millman, K. Jarrod and Mayorov, Nikolay and Nelson, Andrew R. J. and Jones, Eric and Kern, Robert and Larson, Eric and Carey, C J and Polat, {\.I}lhan and Feng, Yu and Moore, Eric W. and {VanderPlas}, Jake and Laxalde, Denis and Perktold, Josef and Cimrman, Robert and Henriksen, Ian and Quintero, E. A. and Harris, Charles R. and Archibald, Anne M. and Ribeiro, Ant{\^o}nio H. and Pedregosa, Fabian and {van Mulbregt}, Paul and {SciPy 1.0 Contributors}},
  title   = {{{SciPy} 1.0: Fundamental Algorithms for Scientific Computing in Python}},
  journal = {Nature Methods},
  year    = {2020},
  volume  = {17},
  pages   = {261--272},
  adsurl  = {https://rdcu.be/b08Wh},
  doi     = {10.1038/s41592-019-0686-2}
}

@misc{scrappage,
  author       = {{Ministry of Transport}},
  title        = {A vehicle scrappage trial for Christchurch and Wellington: May 2009},
  howpublished = {\url{https://www.transport.govt.nz//assets/Uploads/Report/Scrappage-Report-FINAL.pdf}},
  note         = {Accessed 16 Jun. 2021.},
  year         = 2009
}

@article{sklearn,
  title   = {Scikit-learn: Machine Learning in {P}ython},
  author  = {Pedregosa, F. and Varoquaux, G. and Gramfort, A. and Michel, V. and Thirion, B. and Grisel, O. and Blondel, M. and Prettenhofer, P. and Weiss, R. and Dubourg, V. and Vanderplas, J. and Passos, A. and Cournapeau, D. and Brucher, M. and Perrot, M. and Duchesnay, E.},
  journal = {Journal of Machine Learning Research},
  volume  = {12},
  pages   = {2825--2830},
  year    = {2011}
}

@misc{SN21,
  author       = {{Statistics Netherlands}},
  title        = {Quarterly estimates of greenhouse gas emissions},
  howpublished = {\url{https://www.cbs.nl/en-gb/custom/2020/37/quarterly-estimates-of-greenhouse-gas-emissions}},
  note         = {Accessed 8 Nov. 2021.},
  year         = 2021
}

@misc{SNZ_quarterly_emissions_example,
  author       = {{Statistics New Zealand}},
  title        = {Greenhouse gas emissions (industry and household): March 2021 quarter.},
  howpublished = {\url{https://www.stats.govt.nz/experimental/greenhouse-gas-emissions-industry-and-household-march-2021-quarter}},
  note         = {Accessed 8 Oct. 2021.},
  year         = 2021
}

@article{SP95,
  title     = {Bounded-variable least-squares: an algorithm and applications},
  author    = {Stark, Philip B and Parker, Robert L},
  journal   = {Computational Statistics},
  volume    = {10},
  pages     = {129--129},
  year      = {1995},
  publisher = {PHYSICA-VERLAG GMBH}
}

@incollection{Ulk22,
  doi       = {10.1007/978-3-031-09176-6_13},
  url       = {https://doi.org/10.1007/978-3-031-09176-6_13},
  year      = {2022},
  publisher = {Springer International Publishing},
  pages     = {109--116},
  author    = {Ilayda Ulku and Eyup Emre Ulku},
  title     = {Forecasting Greenhouse Gas Emissions Based on Different Machine Learning Algorithms},
  booktitle = {Lecture Notes in Networks and Systems}
}

@article{WEN2020137194,
  doi       = {10.1016/j.scitotenv.2020.137194},
  url       = {https://doi.org/10.1016/j.scitotenv.2020.137194},
  year      = {2020},
  month     = may,
  publisher = {Elsevier {BV}},
  volume    = {718},
  pages     = {137194},
  author    = {Lei Wen and Xiaoyu Yuan},
  title     = {Forecasting {CO}2 emissions in China's commercial department,  through {BP} neural network based on random forest and {PSO}},
  journal   = {Science of The Total Environment}
}

@article{YANG2020124071,
  doi       = {10.1016/j.jclepro.2020.124071},
  url       = {https://doi.org/10.1016/j.jclepro.2020.124071},
  year      = {2020},
  month     = dec,
  publisher = {Elsevier {BV}},
  volume    = {277},
  pages     = {124071},
  author    = {Wenyue Yang and Suhong Zhou},
  title     = {Using decision tree analysis to identify the determinants of residents' {CO}2 emissions from different types of trips: A case study of Guangzhou,  China},
  journal   = {Journal of Cleaner Production}
}

@article{yona2020refining,
  doi       = {10.1007/s13280-019-01312-9},
  url       = {https://doi.org/10.1007/s13280-019-01312-9},
  year      = {2020},
  month     = jan,
  publisher = {Springer Science and Business Media {LLC}},
  volume    = {49},
  number    = {10},
  pages     = {1581--1586},
  author    = {Leehi Yona and Benjamin Cashore and Robert B. Jackson and Jean Ometto and Mark A. Bradford},
  title     = {Refining national greenhouse gas inventories},
  journal   = {Ambio}
}

@article{zheng2020satellite,
  doi       = {10.1126/sciadv.abd4998},
  url       = {https://doi.org/10.1126/sciadv.abd4998},
  year      = {2020},
  month     = dec,
  publisher = {American Association for the Advancement of Science ({AAAS})},
  volume    = {6},
  number    = {49},
  author    = {Bo Zheng and Guannan Geng and Philippe Ciais and Steven J. Davis and Randall V. Martin and Jun Meng and Nana Wu and Frederic Chevallier and Gregoire Broquet and Folkert Boersma and Ronald van der A and Jintai Lin and Dabo Guan and Yu Lei and Kebin He and Qiang Zhang},
  title     = {Satellite-based estimates of decline and rebound in China's {CO}2 emissions during {COVID}-19 pandemic},
  journal   = {Science Advances}
}
    
\end{document}